\documentclass[preprint,11pt]{elsarticle}

\usepackage{amssymb}
\usepackage{algorithm}
\usepackage{algpseudocode}
\usepackage{amsmath}

\graphicspath{ {./figures/} }

\algnewcommand{\Inputs}[1]{%
  \State \textbf{Inputs:}
  \Statex \hspace*{\algorithmicindent}\parbox[t]{.8\linewidth}{\raggedright #1}
}
\algnewcommand{\Initialize}[1]{%
  \State \textbf{Initialize:}
  \Statex \hspace*{\algorithmicindent}\parbox[t]{.8\linewidth}{\raggedright #1}
}

\journal{Artificial Intelligence Journal}

\begin{document}
%\begin{linenumbers}
\begin{frontmatter}



\title{Rapid Open-World Adaptation by Adaptation Principles Learning}


\author[inst1]{Cheng Xue}

\affiliation[inst1]{organization={School of Computing},%Department and Organization
            addressline={The Australian National University}, 
            city={Canberra},
            postcode={2601}, 
            state={ACT},
            country={Australia}}

\author[inst1]{Ekaterina Nikonova}
\author[inst1]{Peng Zhang}
\author[inst1]{Jochen Renz}

\begin{abstract}
Novelty adaptation is the ability of an intelligent agent to adjust its behavior in response to changes in its environment. This is an important characteristic of intelligent agents, as it allows them to continue to function effectively in novel or unexpected situations, but still stands as a critical challenge for deep reinforcement learning (DRL). To tackle this challenge, we propose a simple yet effective novel method, \textit{NAPPING} (\textbf{N}ovelty \textbf{A}daptation \textbf{P}rinci\textbf{p}les Learn\textbf{ing}), that allows trained DRL agents to respond to different classes of novelties in open worlds rapidly. With \textit{NAPPING}, DRL agents can learn to adjust the trained policy only when necessary. They can quickly generalize to similar novel situations without affecting the part of the trained policy that still works. To demonstrate the efficiency and efficacy of NAPPING, we evaluate our method on four action domains that are different in reward structures and the type of task. The domains are CartPole and MountainCar (classic control), CrossRoad (path-finding), and AngryBirds (physical reasoning). We compare \textit{NAPPING} with standard online and fine-tuning DRL methods in CartPole, MountainCar and CrossRoad, and state-of-the-art methods in the more complicated AngryBirds domain. Our evaluation results demonstrate that with our proposed method, DRL agents can rapidly and effectively adjust to a wide range of novel situations across all tested domains.
\end{abstract}


% \begin{highlights}
% \end{highlights}

\begin{keyword}
open world learning \sep novelty adaptation
% %% PACS codes here, in the form: \PACS code \sep code
% %% MSC codes here, in the form: \MSC code \sep code
% %% or \MSC[2008] code \sep code (2000 is the default)
\end{keyword}

\end{frontmatter}

%\linenumbers

\section{Introduction}
\label{sec:introduction}

With the ability to handle complex and noisy inputs, deep reinforcement learning (DRL) \cite{arulkumaran2017brief} has witnessed significant achievements in recent years, achieving super-human performance in multiple domains \cite{Vinyals2019, silver2016mastering, lazaridis2020deep}. Despite the remarkable advancements, DRL algorithms excel in domains often considered closed worlds - where the state space, action space, and transition probability distribution are assumed to remain unchanged. The performance of DRL agents can drop catastrophically when there are minor variations to the environments; even when these changes render the tasks in the environments easier \cite{witty2021measuring}. On the other hand, humans can adjust to novel, unexpected situations efficiently and effectively. More specifically, let us consider the following scenario: \\
\textit{A driver is driving a car downhill and is riding the brakes continually to keep the speed in check. After a while, the driver noticed suddenly that the brake system was not as responsive as before, probably because the brake system heated up, causing a brake fluid leak. From that point, the driver knows that whenever the car needs to be slowed down or stopped, he/she needs to hit the brake earlier.}

In this scenario, the driver was able to quickly adapt to the novel situation of a malfunctioning brake system without affecting the rest of the driving strategy. When considering how humans react to sudden changes, there are three essential characteristics that current DRL methods lack. Firstly, humans adjust the action only when the previously learned action does not work anymore, i.e., humans know that in situations where the novelty does not affect the performance of the task,  adaptation is not required. For example, in the scenario mentioned above, a driver will try to adjust only when the driver wants to slow down the car, e.g., a car in front of the driver travels slower. No changes are required when a car travels in another lane in the opposite direction. While standard deep learning methods, such as online learning and fine-tuning, do not have such guarantees. Learning new policies often causes the agent to forget what has been learned, resulting in the agent re-learning the entire task. Secondly, once humans understand how to adjust to novelties, they do not just adjust in one particular situation but also generalize to other similar situations. It means that when someone understands to hit the brake earlier, the person does not just apply this change of action to one particular situation but also to other similar situations. For example, the driver learns to slow down the car when approaching one intersection with a red traffic light and can generalize to any similar intersections. Lastly, humans can quickly adapt to novel situations, while standard deep learning methods, especially model-free reinforcement learning algorithms, are usually sample-inefficient. These algorithms require a considerable amount of interactions to learn something useful. 

As our response to address the aforementioned challenges, we present \textit{NAPPING} (\textbf{N}ovelty \textbf{A}daptation \textbf{P}rinci\textbf{p}les Learn\textbf{ing}), a novel learning algorithm that works along with deep reinforcement learning agents, which are trained only in pre-novelty environments, to accommodate novel situations when exposed in post-novelty environments quickly. In short, for a trained DRL agent, \textit{NAPPING} identifies regions of states where the previous policy does not work anymore and systematically adjusts the actions in only the identified regions. Ideally, the identified states region should contain conceptually similar states to allow quick generalization; therefore, instead of using the raw state space of an environment, we utilize the embedded representation of states from the trained agent. After that, \textit{NAPPING} learns an \textit{adaptation principle} for each of the regions on-the-fly. The adaptation principles are new policies that can replace the agent's policy in the regions requiring adjustments. 

\begin{figure}[h]
\centering
\includegraphics[width=13cm, height=10cm]{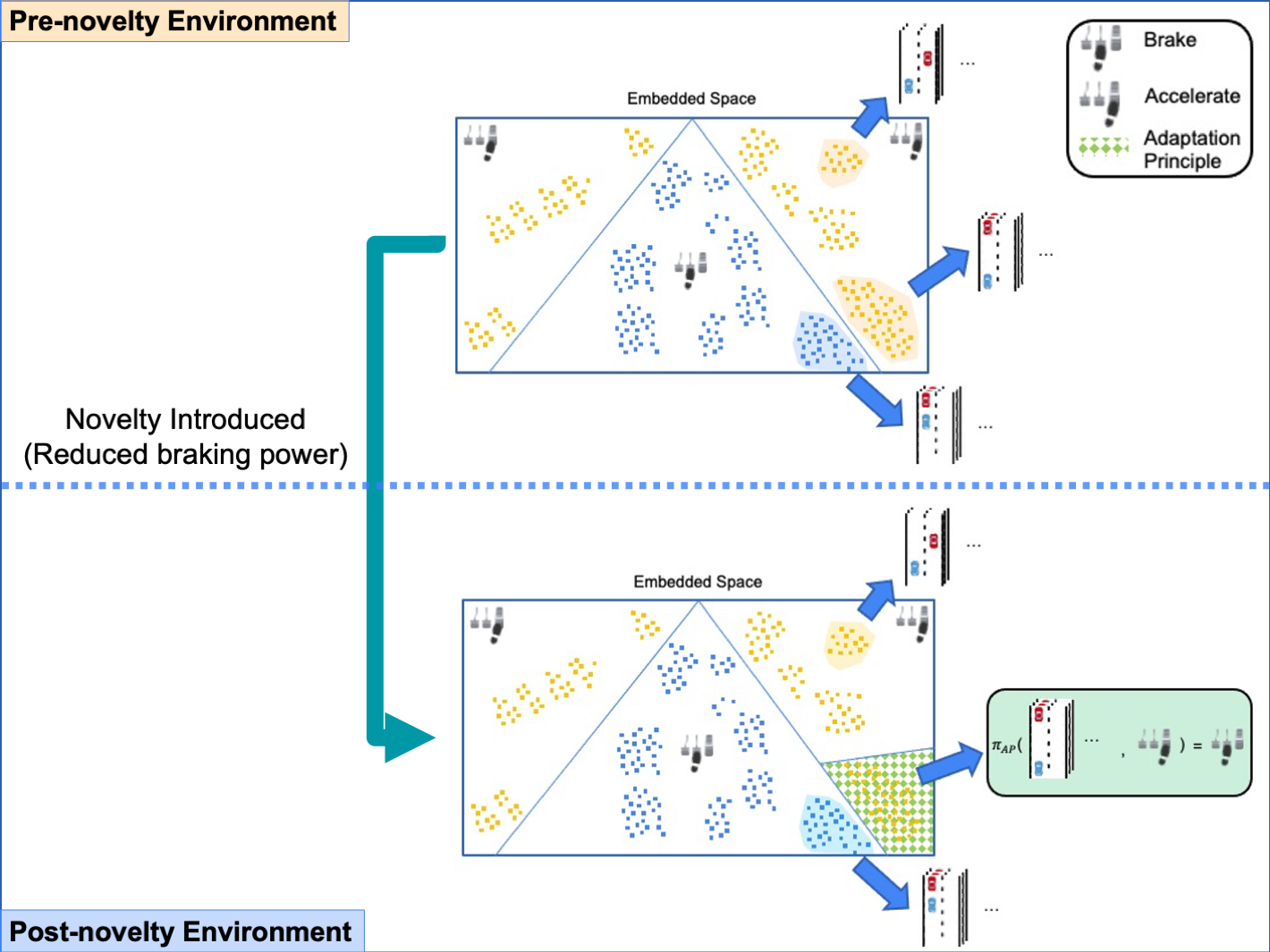}
\caption{Schematic Representation of an \textit{adaptation principle}}
\label{fig:adaptation_principle_example}
\end{figure}

Figure \ref{fig:adaptation_principle_example} is a schematic representation of \textit{NAPPING}. The embedded policy space at the top displays the decision boundary of a trained agent. The yellow points represent states where the agent chooses to hit the accelerator, and the blue points are where the agent hits break. In the pre-novelty environment, the agent hits the accelerator when the car in front of it travels slower but is far away, or there is a car in another lane. The agent will slow down the car when the vehicle in front of the agent car is close to it. After the "Reduced Braking Power" novelty is presented, the agent needs to hit the brake earlier; therefore, the agent needs to learn the \textit{adaptation principle}: when there's a slow car ahead of the agent and it is far away, instead of keeping accelerating, the agent needs to hit the brake instead. The adaptation principle is a way for the agent to adjust its behavior in response to changes in its environment. In this case, the agent needs to hit the brake earlier when there is a slow car in front of it, in order to compensate for the reduced braking power. This allows the agent to continue to navigate safely in its environment, even when faced with unexpected changes. The ability to adapt to new situations is an important characteristic of intelligent agents. We discuss the details of our method in section \ref{sec:adptationprinciples}.

To demonstrate the efficiency and effectiveness of our method, we evaluate \textit{NAPPING} in four different environments. There are two popular control domains CartPole and MountainCar, one path-finding domain CrossRoad, and one physical reasoning domain AngryBirds, using the Science Birds Novelty test-bed \cite{Xue2022}. The performance of our method is compared to the online learning version and the fine-tuning version of the baseline agents in CartPole, MountainCar and CrossRoad and is compared to two state-of-the-art novelty adaptation agents in AngryBirds. For each domain, we create novel situations by varying extensively the physical and spatial parameters of each environment except for the AngryBirds domain, where we evaluate our method on the provided novelties in the 2022 AIBirds Competition Novelty Track \cite{AIBirds}. The evaluation results suggest that while the baselines model fails on almost all novelties, our method rapidly and effectively reacts to novel situations, accommodating many novelties in a few episodes.

\section{Related Work}
\label{sec:relatedwork}
Novelty adaptation is a sub-task in open-world learning, an emerging research area that has received significant attention in recent years. Proposed initially in \cite{Senator2019}, open-world learning has been studied extensively from different theoretical perspectives, ranging from the formal definition of the research problems \cite{Langley2020, Langley2022, https://doi.org/10.48550/arxiv.2011.12906}, the types and definitions of novelties \cite{Langley2020, Boult2021, Molineaux2022}, to various agent frameworks and designs \cite{Langley2020, Langley2022, Molineaux2022, Liu2022AIAS, muhammad2021novelty, burachasmetacognitive} that allow the agent to work in environments with unexpected novelties. The difference between open-world learning and related paradigms, including transfer learning, meta-learning, open-set recognition, incremental learning, zero-shot learning and many others are discussed in \cite{Langley2020, https://doi.org/10.48550/arxiv.2011.12906, goel2022rapid, muhammad2021novelty, NovGrid}.

Existing DRL approaches typically adapt to novelties that gradually change the environments \cite{khetarpal2020towards, padakandla2020reinforcement, cheung2020reinforcement}, while the problem we focus on in this work involves sudden long-term changes to the environment. Meta-reinforcement learning enables AI agents to learn new skills quickly by learning how to learn using prior experience. However, typical meta-reinforcement learning algorithms require AI agents to be trained on multiple distinct tasks to find a nice initialization of weights or hyperparameters before the agents can solve new tasks \cite{yu2020meta}. While in this work, we focus on scenarios where an agent is trained on only one specification of the environment and there exist baseline agents with acceptable performance for the pre-novelty environment \cite{Langley2020}. Closely related to novelty adaptation in open worlds, transfer reinforcement learning agents often need to 1) select one or some appropriate source tasks given a target task, 2) learn how the source tasks and target task are related, and finally, 3) transfer the knowledge from the source task to the target task \cite{taylor2009transfer}. The closest analogy to our proposed method, transfer learning via policy reusing aims to reuse policies learned from source tasks to construct the policy in the target task \cite{zhu2020transfer}. While we also reuse some parts of the policy from a trained agent, we partition the trained policy into multiple regions, reuse the regions that still work, and relearn those that do not work as expected. Transfer learning via policy reuse approaches, on the other hand, typically learns multiple policies or Q-value functions in the source task, and the action of the agent is chosen from the policy that gives the highest Q-value for a given state for the target task \cite{fernandez2006probabilistic, barreto2017successor}. 

Several approaches have recently been developed to address the open-world learning challenge, with some mainly focusing on novelty detection \cite{mclure2022changepoint,boult2022weibull,li2021unsupervised} and others on novelty adaptation \cite{goel2022rapid, kumar2021rma,sternmodel,klenk2020model, musliner2021openmind}. Among the novelty adaptation approaches, \cite{kumar2021rma} introduced RMA (Rapid Motor Adaptation), allowing legged robots to adapt to changing terrains, payloads, wear, and tear. Similar to ours, their method consists of a base policy and an adaptation module. However, their base policy needs to be trained in a simulator with different environmental parameters, e.g., robot mass, friction, terrain height, and motor strength. In other words, their method handles known unknowns while \textit{NAPPING} does not assume the kind of novel situations it may encounter; hence \textit{NAPPING} is capable of handling unknown unknowns. HYDRA \cite{klenk2020model, sternmodel} is a planning-centric agent that can detect and adapt to novelties in open worlds by using a compositions model of the environment defined using PDDL+, and a range of AI techniques, including domain-independent automated planning, anomaly detection, model-based diagnosis, and heuristic search. OpenMIND \cite{musliner2021openmind} is similar to HYDRA but with a different planning formalism. It consists of a goal-oriented agent that plans to achieve its goals, executes its plans and revises its goals and planning models on the fly when novel, unexpected situations arise. Similar to \cite{klenk2020model, sternmodel}, \cite{musliner2021openmind} also introduces several heuristics that have proven effective in handling novelty. A hybrid planning and learning method RAPid-Learn \cite{goel2022rapid} has been recently proposed to address the same problem with domain knowledge grounded using PDDL and is evaluated on a grid world environment inspired by Minecraft. For RAPid-Learn, the agent needs to know the set of known entities, a set of known predicates, a set of symbolic states in the environment, and a set of known actions and their preconditions and effects. The main difference between our method and HYDRA, OpenMIND and RAPid-Learn is that our method does not assume explicit understanding of the structure of the environments, including but not limit to a model of the environment, the possible set of novelties, handcrafted predicates and preconditions and results of available actions. Also, our method is not based on planning or a formal language e.g. PDDL.

\section{Problem Definition}
\label{sec:problemdefinition}
Similar to \cite{klenk2020model, sternmodel} we assume agents play a sequence of episodes, every episode has a sequence of steps consisting of a state, an action, and a subsequent state $(s,a,s')$. The environment is defined as a Markov Decision Process (MDP) $E = <S, A, Pr, R>$, where $S$ is the set of environment states, $A$ is the set of available actions of the agent, $Pr(s_{t+1} | a_{t}, s_{t})$ is the transition probability distribution that returns the probability of $s_{t+1}$ given $a_t$ and $s_t$, and $R$ is the reward function that returns the immediate reward after the transition from $s_t$ to $s_{t+1}$ with action $a_t$. In each episode, the agent receives a state $s_t \in S$  selects an action $a_t \in A$, and then receives the reward $r_{t+1}$ for the action $a_t$ and the resulting state $s_{t+1}$ until the episode ends. 

Following the \textit{Theory of Environment Change} of \cite{Langley2020}, we consider novelties are transformations of the underlying environment $E$. More explicitly, a novelty is defined as a transformation function $\phi$; a post-novelty environment is represented as $\phi(E) = <S_{\phi}, A_{\phi}, P_{\phi}, R_{\phi}>$. In other words, a novelty may or may not change the state space, action space, transition probability, and the reward function of the environment.

\section{\textit{NAPPING} - \textbf{N}ovelty \textbf{A}daptation \textbf{P}rinci\textbf{p}les Learn\textbf{ing} }
\label{sec:adptationprinciples}

\textit{NAPPING} accommodates novelties by first identifying the regions in the embedded state space where a trained DRL agent under-performs and then varying the action for each individual region, aiming to learn a better action for each such region. We call each region and its corresponding action an \textit{\textit{adaptation principle}}. The reason is that each \textit{adaptation principle} can be loosely explained in terms of a rule: "If the agent is in some situation $s$ and wanted to do $x$, it should now do $y$ instead". Formally, we define the $i$th \textit{adaptation principle} to be a function that is defined over the $i$th model embedded state set $MS^{i}_{ap}$ as $\pi^i_{ap}:MS^{i}_{ap} \rightarrow A$ where $A$ is the set of the available actions of the agent. Therefore, assuming there are $N$ \textit{adaptation principles}, the NAPPING policy is,

\begin{equation*}
\pi^{NAPPING}_t(s_t, ms_t) = 
\begin{cases}
  \pi^{baseline}(s_t) \text{ if } ms_t \notin \{MS^{i}_{ap}\}_{i=1}^N \\\
  \pi^{i}_{ap}(ms_t) \text{ if } ms_t \in MS^{i}_{ap}
\end{cases}
\end{equation*}
We further refer to the \textit{adaptation principle} that is still learning an \textit{open \textit{adaptation principle}} and a learned one a \textit{closed \textit{adaptation principle}}. In the following, we discuss how we identify the region of model state $MS^{i}_{ap}$ for \textit{\textit{adaptation principle}} $\pi^i_{ap}$ and how to learn the mapping of $\pi^i_{ap}$ to the action space.

\begin{figure}[h]
\centering
\includegraphics[width=16cm, height=8cm]{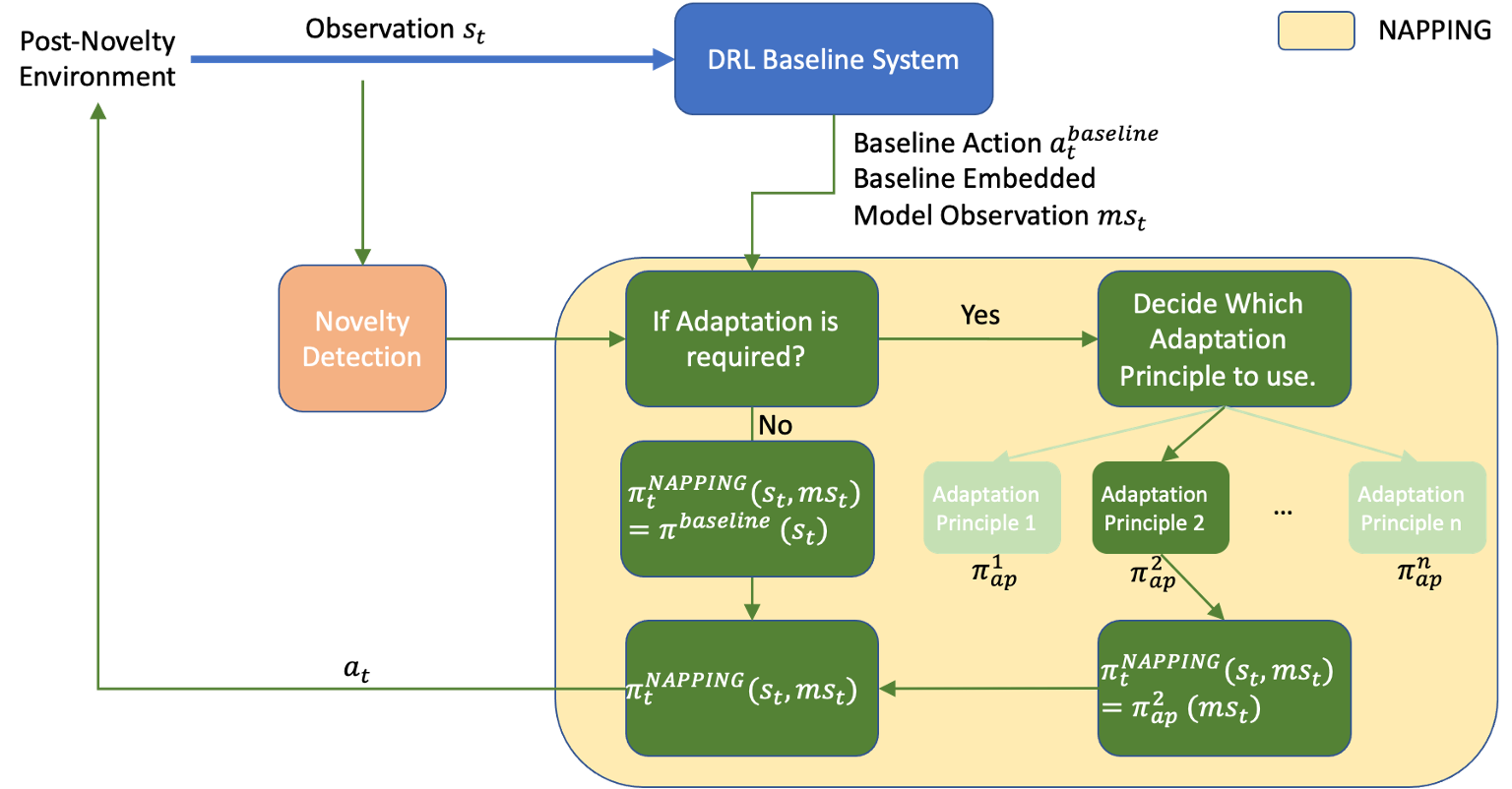}
\caption{High-level Architecture of NAPPING}
\label{fig:learning_adaptation_principle_example}
\end{figure}

We now discuss in detail how an adaptation principle region is identified and how an adaptation principle is learned, followed by an example of \textit{NAPPING}.

\paragraph{\textbf{Identifying adaptation principle regions}} Consider a DRL agent with a neural network policy where the last layer outputs a $n$-dimensional vector representing the action in the $n$-dimensional action space, we assume that the neural network, independent of the architecture (e.g., CNN, FC, RNN, or Transformer) should embed raw environmental states to a space that can be linearly separated to allow the last output layer to select the desired action. Thus, semantically similar states should be close to each other in the embedded space. Based on this assumption, and to improve the efficiency of \textit{NAPPING}, we believe an adaption principle, if required, should affect not only the model state $ms_t$ but also its "surrounding area" in the embedded space. Subsequently, the question is how one can define the exact boundary of the "surrounding area" a principle should cover. A naive approach is partitioning the embedded space uniformly along each dimension into cells. However, this approach requires a relatively large number of model states $ms_t$ to cover enough cells to produce reliable predictions. However, since we would like the agent to adapt to novel situations as fast as possible, the number of environment states available to the agent to produce $ms_t$ will be limited. Moreover, the number of cells required to achieve reasonable precision increases exponentially along with the dimension of the embedding space; uniformly partitioning the space is not an effective method and can quickly become intractable. Instead, we create Voronoi diagrams \cite{aurenhammer1991voronoi} in an adaptive manner to divide the embedded space. A Voronoi diagram is a partition of a plane into regions close to each of a given set of points. For each point, there is a corresponding region, a Voronoi cell, which consists of all points of the plane closer to that point than to any other. Therefore, each $MS^{i}_{ap}$ is a Voronoi cell created by a $ms_t$. Next, we must decide what model states will be used to create the Voronoi cells, hence later, the adaptation principles. Each \textit{adaptation principle} must learn the corresponding action independently; therefore, if we include all the states that the agent encounters in the novel environment, we may have too many principles to learn, hence reducing the method's performance. As a solution in this work, we define a simple function $Eval:S \times A \times S \rightarrow \mathbb{R}$ that assigns a real-valued score to the action that \textit{NAPPING} performs and we define a threshold $thre$ for each domain. We include $ms_t$ incrementally to create a new Voronoi diagram when the $Eval(s_t, a_t, s_{t+1})$ returns a value less than the threshold. For example, in the physical reasoning domain Angry Birds, the function $Eval(\cdot)$ returns the reward of an action, which is the number of pigs that were destroyed. The threshold \textit{thre} for Angry Birds is set to 1. It means that $ms_t$ is included among the existing model states to create a new Voronoi diagram if the action $a_t$ has not destroyed any pig. It is worth noting that although we manually define the function and threshold in this work, future work may focus on using different methods, for instance, a trained Q-value function, to serve the same purpose. 

\paragraph{\textbf{Learning adaptation principles}} We now discuss how \textit{NAPPING} learns an \textit{adaptation principle}. Once \textit{NAPPING} decides an \textit{adaptation principle} is required for a model state $ms_t$, a new Voronoi cell $MS^{i}_{ap}$ is generated. For $MS^{i}_{ap}$, a set of available actions is then initialized from the action space $A$, and the \textit{adaptation principle} becomes an \textit{open \textit{adaptation principle}}. We call this set of actions the candidate actions set $A^i_{ap}$ for $\pi^{i}_{ap}$. Assuming a model state $ms_t$ falls in the region $MS^{i}_{ap}$, the agent chooses an action $a^{ap}_{t}$ from $A^i_{ap}$ and evaluates using $Eval(\cdot)$. We also initialize a dictionary $BestScore$ to store the best action values of each $ms_t$ used to create the Voronoi diagram. To decide if $\pi^{i}_{ap}(ms_t) = a^{ap}_{t}$ can replace $\pi_{baseline}$ in $MS^{i}_{ap}$, we compare the action value of $a^{ap}_{t}$ against both the $thre$ and $BestScore$. If the value of $a^{ap}_t$ is greater than or equals to both the value of $a_{baseline}$ and $thre$, $a^{ap}_{t}$ is considered a good action and the value of $a^{ap}_{t}$ updates the $BestScore$ for $ms_t$. Otherwise, the $a^{ap}_{t}$ is removed from the available action set, and the agent will try the rest of the actions. The \textit{adaptation principle} becomes a \textit{closed \textit{adaptation principle}} when only one action is in the available action set. In some cases, the $Eval(s_t, a^{ap}_t, s_{t+1})$ may reach the highest value an action can get. For those cases, the \textit{adaptation principle} is closed directly with just $a^{ap}_t$ as no other action will evaluate to a higher value. In practice, we use the algorithms in \ref{sec:algorithm:policy} and in \ref{sec:algorithm:update} to realize our method.

\begin{figure}[h]
\centering
\includegraphics[width=16cm, height=12cm]{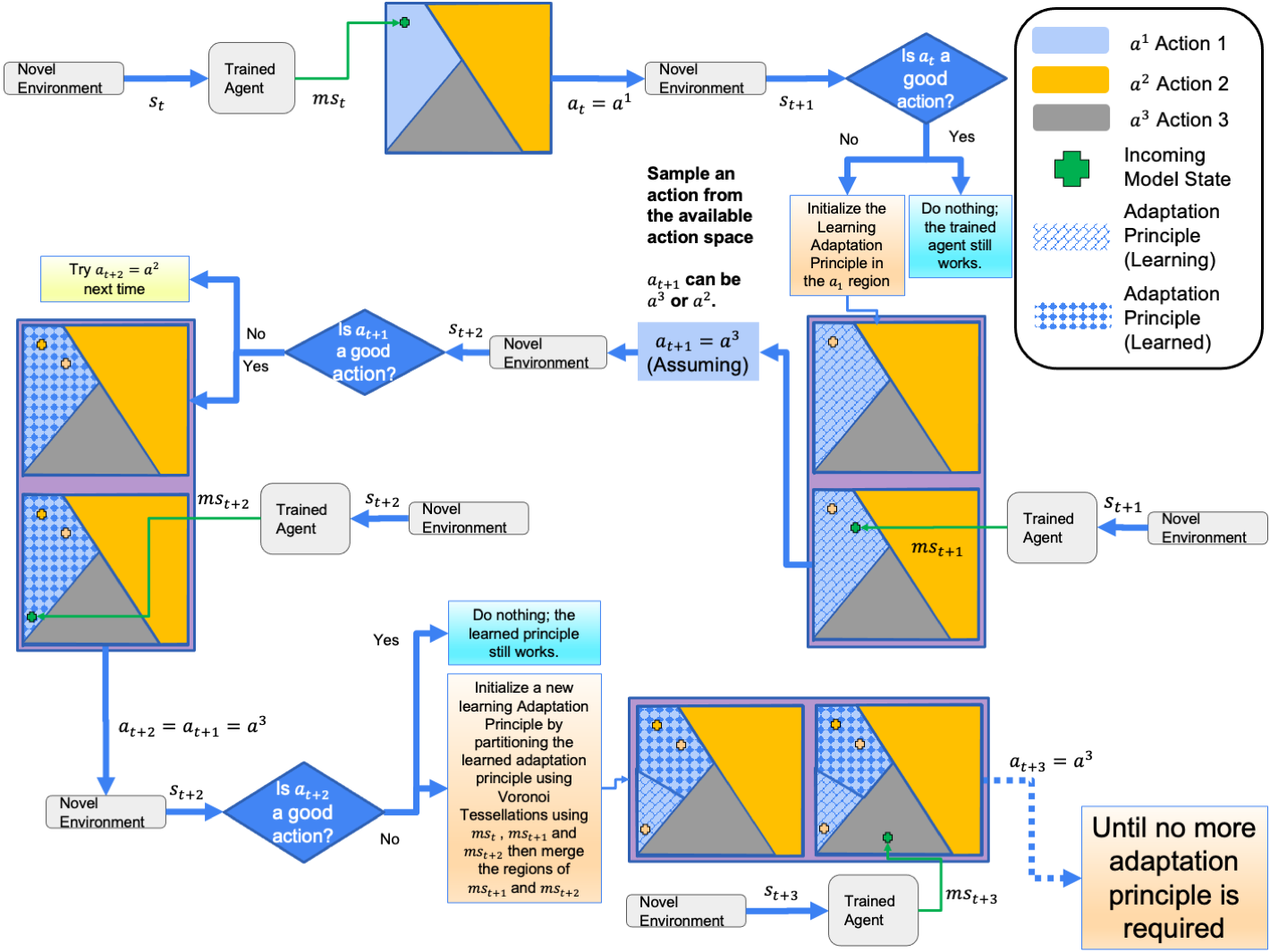}
\caption{An Example of How to Learn Adaptation Principles}
\label{fig:learning_adaptation_principle_example}
\end{figure}

\paragraph{\textbf{An Example of \textit{NAPPING}}} Fig \ref{fig:learning_adaptation_principle_example} is an exemplar flowchart of how adaptation principles are learned. Assuming a trained agent with $3$ available actions $a^1, a^2,$ and $a^3$, represented by blue, yellow, and grey regions, respectively, in the trained agent's embedding space. Once a novelty is introduced, the trained agent processes the post-novelty environment state $s_t$ to the model state $ms_t$. As this is the first model state, there is no associated \textit{adaptation principle}. Hence, as the model state falls in the blue region, $\pi^{NAPPING}_t(s_t, ms_t)=\pi^{baseline}(s_t)=a^1$. After the agent stepped $a_t=a^1$ the action and obtains the next environment state $s_{t+1}$, the agent evaluates if $a_t$ is a good action by comparing the $Eval(s_t,a_t,s_{t+1})$ with the predefined threshold and the best scores dictionary. If $a_t$ is a good action, we can ignore it as the trained agent still works in $ms_t$. Otherwise, an \textit{adaptation principle} is required, and the region $MS^{1}_{ap}$ is initialized to cover the whole space of $a^1$. It is essentially assuming that now $a^1$ does not work anymore at all. When the \textit{adaptation principle} region is just initialized, the \textit{adaptation principle} becomes an \textit{open \textit{adaptation principle}}. The corresponding action space is initialized to contain all possible actions as candidate actions for the principle, except the action that the baseline agent has already selected, e.g., the candidate action set is $\{a^2, a^3\}$. The agent will sample one action from the candidate action space each time the \textit{adaptation principle} region is activated and will remove the action from the candidate action space if the sampled action is not good. %The \textit{adaptation principle} is closed when only one action with the highest $Eval$ value exists in the \textit{adaptation principle}'s candidate action space. It means that if any future model state falls in the \textit{adaptation principle} region, the only action in the candidate action space of the \textit{adaptation principle} is used.
In Fig \ref{fig:learning_adaptation_principle_example}, as the \textit{adaptation principle} is open at $t+1$, we assume the action sampled at $t+1$ is action 3, resulting $a_{t+1} = a^3$. After getting the next state $s_{t+2}$, the adaptation agent evaluates if $a_{t+1}$ is a good action. Suppose $a_{t+1}$ is good and $Eval(s_{t+1}, a^3, s_{t+2})$ reaches the maximum possible value. In that case, the state of the \textit{adaptation principle} becomes closed as the algorithm stops searching for other actions because the $a_{t+1}$ is already the best possible action. Otherwise, the adaptation agent will keep trying other available actions when a model state falls in the \textit{adaptation principle} region. Following the same logic, let us assume the next state $s_{t+2}$ leads to a model state $ms_{t+2}$ that still falls in the learned \textit{adaptation principle} region but is considered an undesired action. In such a situation, we partition the \textit{adaptation principle} region using Voronoi Tessellation using $ms_t$ and $ms_{t+2}$. So, now we have a new \textit{adaptation principle} that is learning (open), and we keep the learned (closed) \textit{adaptation principle} as it is. We restate that the reason for using Voronoi Tessellation to partition the embedded space is that conceptually related states should be close to each other in the embedded space of a trained agent. When we do not have enough model state $ms$ to decide the exact boundary of regions that require changing policies, a sensible way is to assume that states close to the existing model state should have the same \textit{adaptation principle}. The agent uses the \textit{adaptation principle} until the principle fails to work. That is when a new \textit{adaptation principle} region is created around the model state that the old principle fails. The algorithm continues until a trial terminates or no more \textit{adaptation principle} is required. 

To sum up, our proposed method creates adaptation principles in regions where a trained agent does not perform acceptably. Based on the assumption that the embedded space captures the agent's abstract understanding of the environment, we utilize Voronoi Tessellation to partition the space into regions where the result of the proposed action falls below expectation. Then we search for better action (the adaptation action) for each \textit{adaptation principle} region.

We evaluate our proposed method extensively on four domains with a large range of novel situations. Our results demonstrate that the proposed method achieves rapid adaptation in all domains we tested.

\begin{figure*}
\centering
\includegraphics[width=3cm]{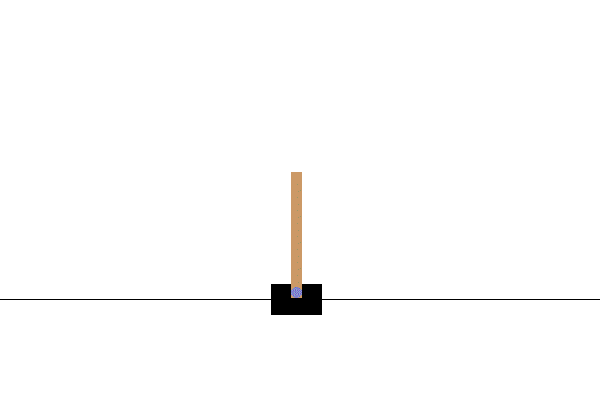}
\includegraphics[width=3cm]{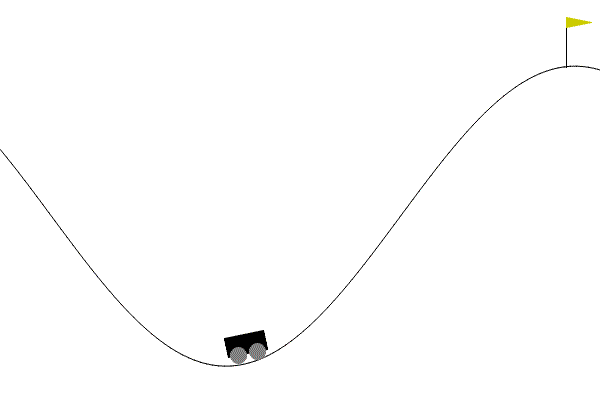}
\includegraphics[width=2cm]{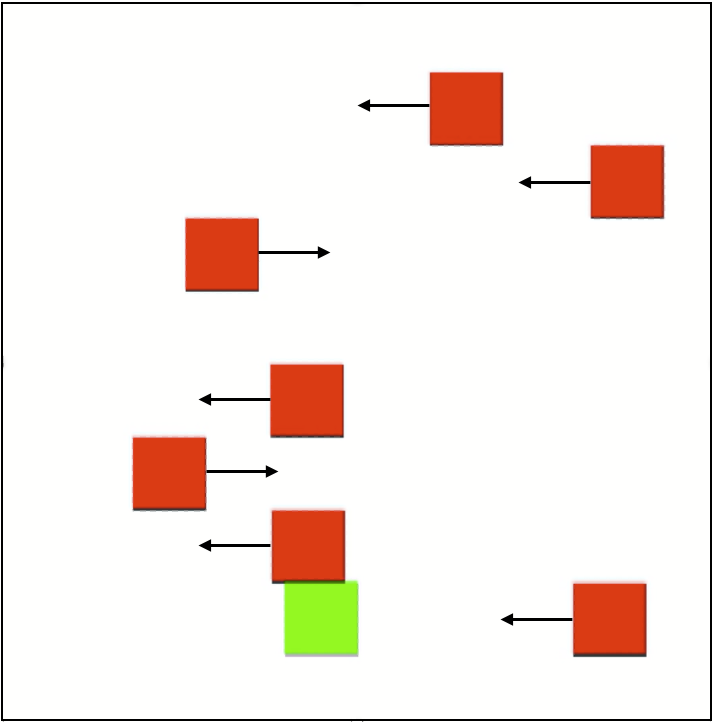}
\includegraphics[width=3cm]{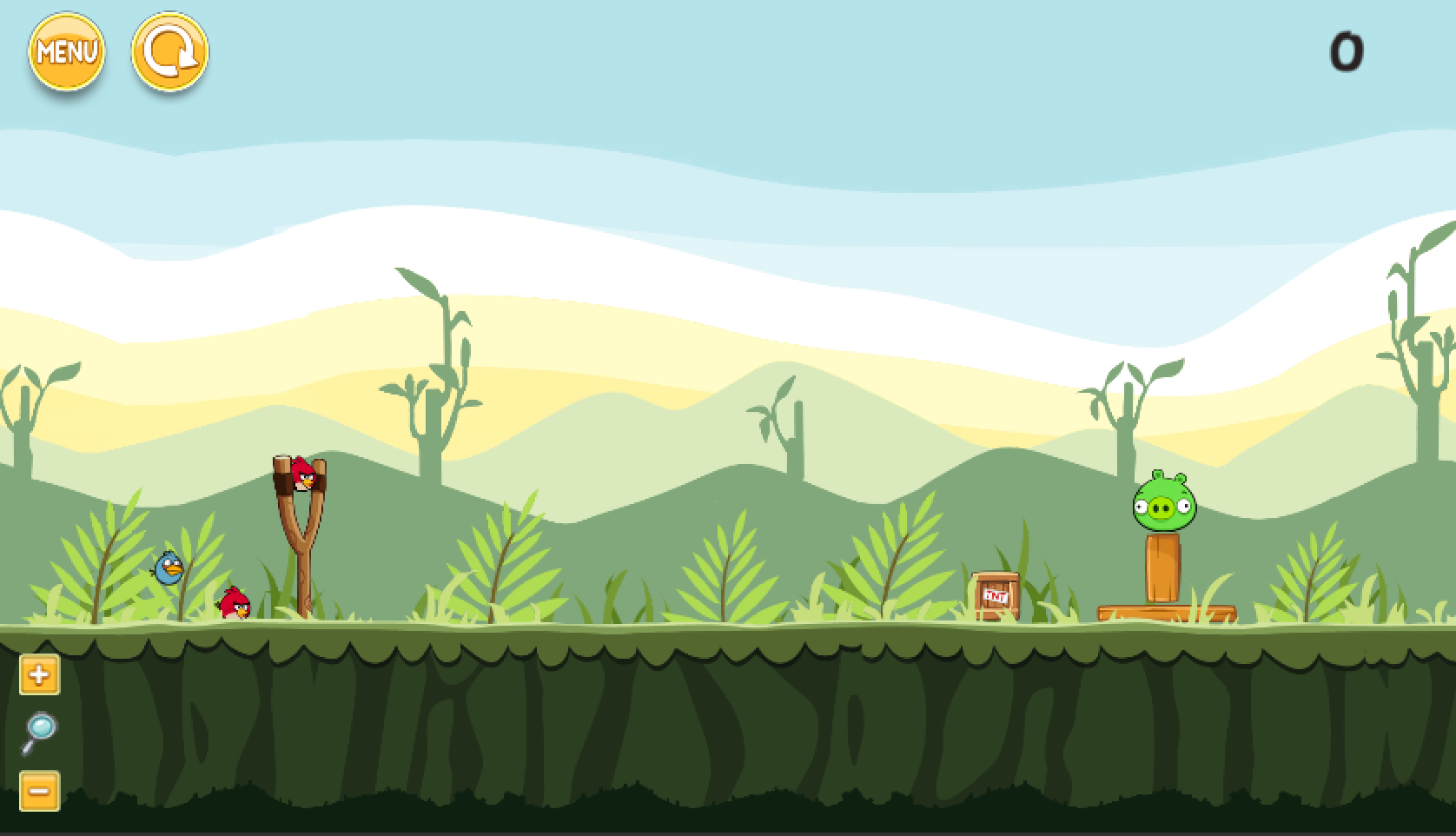}
\caption{ Domains used in this work. From left to right: CartPole, MountainCar, Crossroad, and Angry Birds}
\label{fig:test-domains}
\end{figure*}

\section{Evaluation Results}
\label{sec:results}

\textbf{Domains and Agents} In this work, we test our proposed method on four distinct domains (fig. \ref{fig:test-domains}), namely CartPole \cite{Barto1983NeuronlikeAE}, MountainCar \cite{Brockman2016OpenAIG, Moore90efficientmemory-based}, CrossRoad \cite{CrossroadDomain}, and Angry Birds using Science Birds Novelty \cite{Xue2022}.

\textbf{Evaluation Protocol} We follow the standard open-world learning trial setting discussed in \cite{Pinto2022} and used in \cite{klenk2020model, sternmodel, goel2022rapid, Xue2022} to evaluate our method. A trial consists of a sequence of episodes, which starts from episodes drawn from the pre-novelty environment $E$. After several pre-novelty episodes, a novelty is introduced, and the environment $E$ is transferred to the post-novelty environment $\phi(E)$. After that, all the remaining episodes are drawn from $\phi(E)$. In each trial, there is one and only one transformation applied to the pre-novelty environment. The agent's objective is to maximize the cumulative reward in the trial. In this work, a trial consists of 80 episodes. The first 40 episodes are sampled from the pre-novelty environment, and the other 40 are from the post-novelty environment. The novelty is always introduced in episode 0. Therefore all the pre-novelty episodes have a negative index from -40 to -1. We use the same setting for all domains, except for Angry Birds where the number of pre-novelty episodes in a trial is a random number.

\textbf{Novelty Detection} While novelty detection is an indispensable component in open-world learning, in this paper, we put our focus mainly on novelty adaptation. Therefore, we adopt a simple heuristic in each domain to determine if the novelty has been introduced. That is, the agent reports novelty when the 5-episode rolling average reward is less than a domain-dependent threshold. The novelty detection heuristic is activated from the episode index 5 (the $45$th episode) in every trial in CartPole, MountainCar and CrossRoad. The novelty status is provided to the agent in AngryBirds.

\textbf{How novelties are generated?}
For CartPole, Mountain Car and CrossRoad, post-novelty environments are created by varying the environment's physical or spatial parameter values. We use the novelties presented in the 2022 AIBirds Competition \cite{AIBirds} Novelty Track for the Angry Birds domain. The details are discussed in the following sections.

\subsection{CartPole}
OpenAI Cartpole environment is a classical RL benchmark based on the cart-pole problem described by \cite{Barto1983NeuronlikeAE}. In this environment, a pole is connected to a cart. The task is to maintain the pole upright by moving the cart either left or right. The reward of +1 is given for each step that the pole remains upright. The episode ends if the pole is greater than 12 degrees from the vertical position or if the cart is greater than 2.4 units from the center (i.e. the cart reaches the edge of the display). In this paper, we truncate the maximum possible reward at 200, so the total reward for an episode is between 1 and 200.

\subsubsection{Baseline Agents}
We evaluate our method using a pre-trained DQN and a pre-trained PPO from \cite{stable-baselines3}. The two agents achieve perfect performance in the pre-novel environment. For each agent, we evaluate the performance on the trials with four settings: no learning (-Baseline), online learning (-Online), fine-tuning (-FineTune), and Principle Adaptation(-NAPPING(ours)). 

\subsubsection{Novelty}
We varied the parameters in CartPole to create novel situations for the agents. The parameters that are varied in CartPole are: 1) length of the pole, 2) gravity, 3) mass of the cart, 4) mass of the pole, and 5) the magnitude of the pushing force, each with the default value of 0.5, 9.8, 1, 0.1 and 10 respectively.

To explore the limits of our proposed method, we sample the value for these parameters from an extensive range; we set the lower range to be the default parameter value divided by ten and the upper to be the default value multiplied by 10 of the default value for all parameters. Therefore, the ranges of the parameters are:

\begin{itemize}
\item Length: the length of the pole from 0.05 to 5 unit
\item Gravity: gravity range from 0.98 to 98 unit
\item MassCart: the mass of the cart from 0.1 to 10 unit
\item Masspole: the mass of the pole from 0.01 to 1 unit
\item Force\_mag: the magnitude of the pushing force from 1 to 100 unit
\end{itemize}

For each trial, the novel environment's parameter values are sampled uniformly from the parameter range given above. We run 5000 trials per agent. All agents use the same detection heuristic: a novelty is detected if the five-rolling average episodic reward is less than 150.

\subsubsection{CartPole Results and Discussion}

\begin{figure}[h]
\centering
\includegraphics[width=8cm, height=7cm]{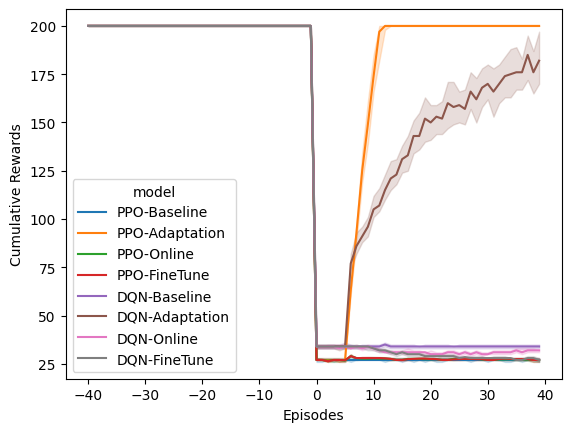}
\caption{Overall Cumulative Rewards on CartPole by Agents}
\label{fig:cartpole_overall}
\end{figure}

Figure \ref{fig:cartpole_overall} shows the overall cumulative rewards on CartPole for all the models we experimented with. We used median instead of mean here to alleviate the impact of the results from unsolvable/challenging novel environments. From the figure, we notice that none of the baseline, online learning, and fine-tuning agents showed any adaptation. In contrast, the PPO-NAPPING agent and the DQN-NAPPING reacted quickly and effectively to the novelties. Significantly, the median cumulative rewards of PPO-NAPPING reach 200 in less than ten episodes. DQN-NAPPING, although showing substantial adaptation progress, adapts less efficiently when compared to PPO-NAPPING. It suggests the PPO learned a more suitable embedding space that allows NAPPING to identify practical adaptation principles quickly.

\begin{figure}[h]
\centering
\includegraphics[width=14cm, height=7cm]{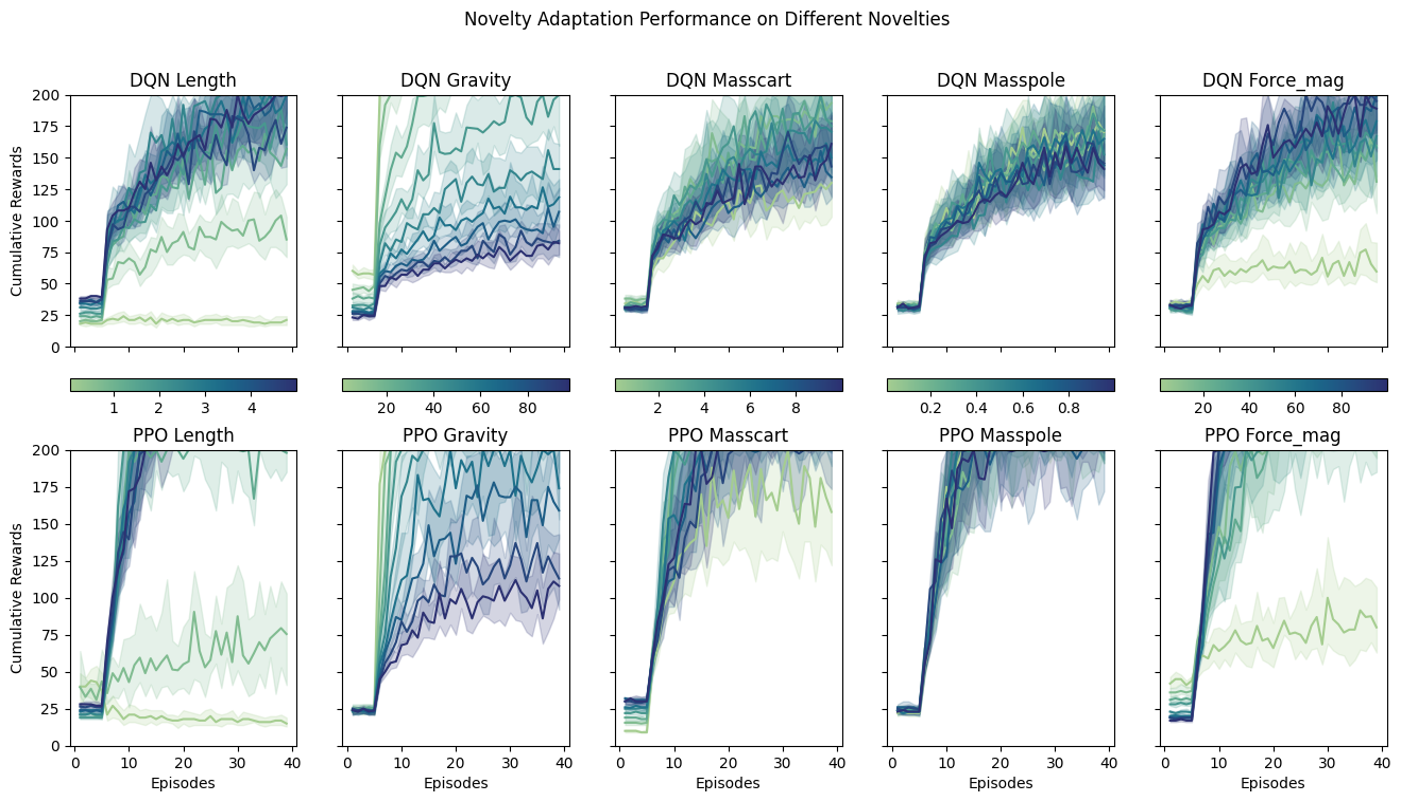}
\caption{Overall Cumulative Rewards on CartPole By Environment Parameter Values}
\label{fig:cartpole_pernovelty}
\end{figure}

Figure \ref{fig:cartpole_pernovelty} demonstrates the DQN-NAPPING and PPO-NAPPING performances by different parameter values for only the post-novelty episodes. As we start to detect novelties from episode 5, the first five episodes illustrate the performance of the DQN-Baseline and PPO-Baseline. Confirming what was discussed in \cite{witty2021measuring}, the performance of both baseline agents dropped significantly even when the task became more manageable, e.g., when the gravity became lower, and the pushing force became larger. As shown in the figures,  \textit{NAPPING} enables standard DQN and PPO to adapt to almost all the novelties, except those adamant ones, e.g., when the length of the pole or the pushing force becomes too small, and when the gravity becomes too large. 

\subsection{MountainCar}
OpenAI MountainCar is a popular control domain that is based on the deterministic MDP problem described in \cite{Moore90efficientmemory-based}. In the domain, an underpowered car moves along a curve and attempts to reach a goal state at the top of the "mountain" by selecting between three actions on every time step, Forward, Stay, and Backward. The Forward action accelerates the car in the positive $x$ direction. Backward accelerates the car in the negative $x$ direction. The agent's state is described by two state variables: the horizontal position, $x$, and velocity, $\dot{x}$. The agent receives a reward of $-1$ per time step. Thus the goal of the agent is to reach the goal state as soon as possible. An episode ends when the agent reaches the goal state or $500$ time steps has passed, whichever comes first. Therefore, the reward for an episode is between 0 to -500.

\subsubsection{Baseline Agents}
In order to examine \textit{NAPPING}'s performance when the baseline agents have different performance levels, we include two versions of PPO, namely the PPO-Weak and PPO-Strong. We also include the DQN agent from \cite{stable-baselines3} for evaluation. PPO-Strong and DQN solve the pre-novelty environment with an average episodic reward of around 120 and -100, respectively. PPO-Weak fails in the pre-novelty environment with -500 average rewards but still manages to work in some less demanding post-novelty environments. PPO-Strong has a superior generalization ability, with solving even the most novel situations. We include this agent to serve as an empirical maximum performance for this domain so that we can evaluate how much NAPPING can help DQN and PPO-Weak to adapt more intuitively.  

\subsubsection{Novelty}
To create novelty situations in MountainCar, we varied the pushing force (default:$0.001$) and the gravity (default:$0.0025$) of the environment. As extreme values in MountainCar can quickly render the game unsolvable, e.g., small pushing force and massive gravity, we select the values of pushing force from $[0.0001,0.02]$ and gravity from $[0.0001, 0.005]$ to create novelties.

Like CartPole, each trial in MountainCar consists of 80 episodes. The first 40 episodes come from the pre-novelty environment, and the rest 40 are from the novel environment. For each trial, the novel environment's parameter values are sampled uniformly from the parameter listed range. We run 2000 trials per agent. Like the agents in CartPole, all agents detect novelty from episode 5 (the $45$th episode) using the same detection heuristic: a novelty is detected if the five-rolling average episodic reward is less than -120 or greater than -80.
\subsubsection{MountainCar Results and Discussion}

\begin{figure}[h]
\centering
\includegraphics[width=15cm, height=5cm]{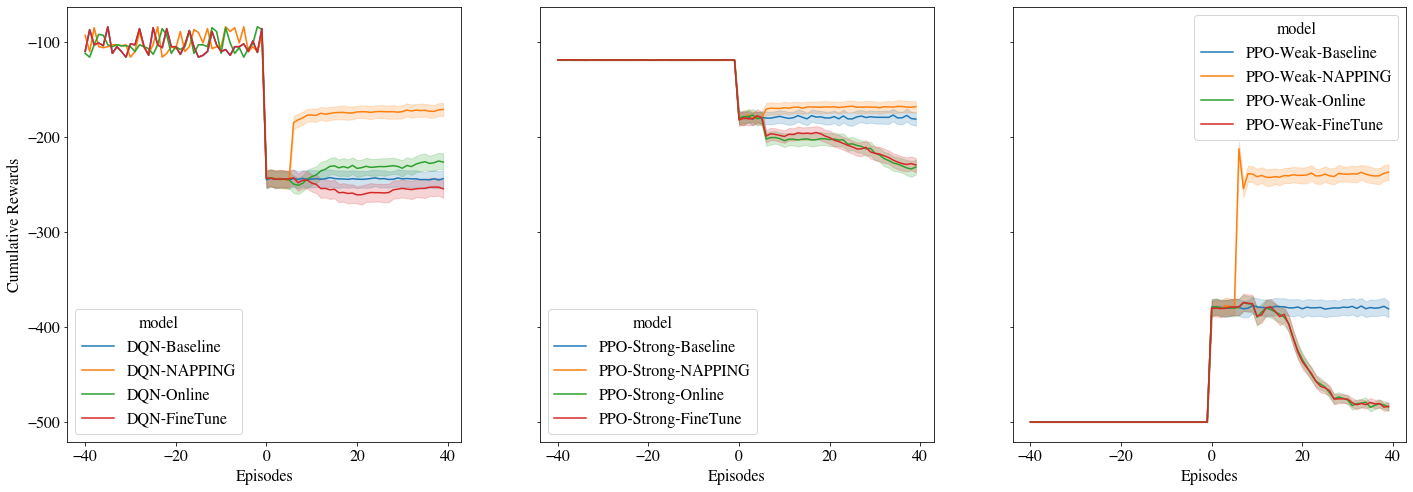}
\caption{Overall Cumulative Rewards on Mountain Car by Agents}
\label{fig:mountaincar_overall}
\end{figure}

Figure \ref{fig:mountaincar_overall} displays the overall cumulative rewards on the Mountain Car domain. It is worth noting that, unlike CartPole, the cumulative rewards in MountainCar reflect how fast an agent reaches the goal state; therefore, it is not always possible for the agent to return to the pre-novelty performance level when the environment becomes harder to solve. It is clear from the figure that all -NAPPING agents adapt to novel situations rapidly, with a significant upward jump in the performance in episode 5 when the agents learn adaptation principles. It suggests that in the MountainCar domain, NAPPING achieves significant adaptation performance within the very first episode in the post-novelty environment. It is also interesting to notice that although DQN performs better in the pre-novelty environment, it has lower average rewards when a novelty is introduced compared to the PPO-Strong agent. While PPO-Strong-NAPPING seems more robust than DQN-NAPPING, they converge quickly to the same performance level. The performance of DQN-NAPPING, although it starts much lower than PPO-Strong, quickly picks up after episode 5 and converges to the same level as PPO-Strong's performance. This indicates that with \textit{NAPPING}, DQN is able to reach the empirical maximum performance level. Although the  PPO-Weak agent fails in the pre-novelty environment, the baseline still solves some of the novelties in MountainCar. The PPO-Weak-NAPPING result shows that even with a weak baseline agent, our method still manages to help the agent react to unknown changes with a significant jump in the performance in the first novel episode. However, we also notice that the asymptotic performance of PPO-Weak is lower than that of PPO-Strong-NAPPING and DQN-NAPPING, indicating that the baseline performance affects the performance of our proposed method.

\begin{figure}[h]
\centering
\includegraphics[width=14cm, height=7cm]{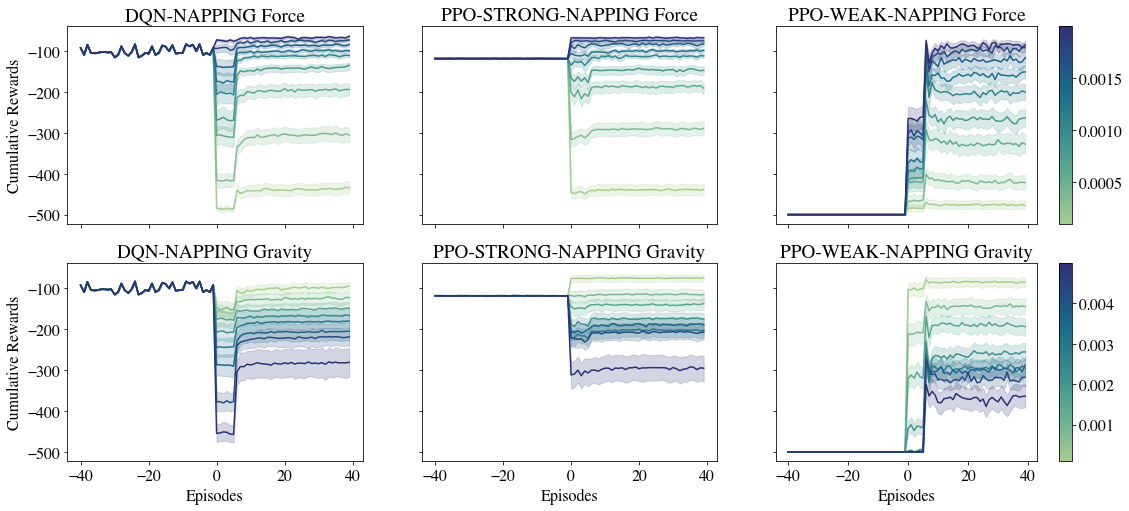}
\caption{Overall Cumulative Rewards on MountainCar by Novelty}
\label{fig:mountaincar_bynovelty}
\end{figure}

Figure \ref{fig:mountaincar_bynovelty} displays the average performance of -NAPPING agents on the Mountain Car domain by different novel parameter values. The first pattern to notice is that unlike PPO-Strong, where the performance seems more consistent with the task's difficulty, DQN-NAPPING's performance dropped across all the gravity values. On the other hand, PPO-Strong achieves better performance when the environment becomes more straightforward (e.g.,  larger pushing force or lower gravity values). Secondly, it is interesting to see that NAPPING helps DRL agents to adapt to not only the novelties that make the task harder (post-novelty baseline performance lower than per-novelty), but also those that make the environment easier (post-novelty baseline performance higher than per-novelty). 

\begin{figure}[h]
\centering
\includegraphics[width=14cm, height=7cm]{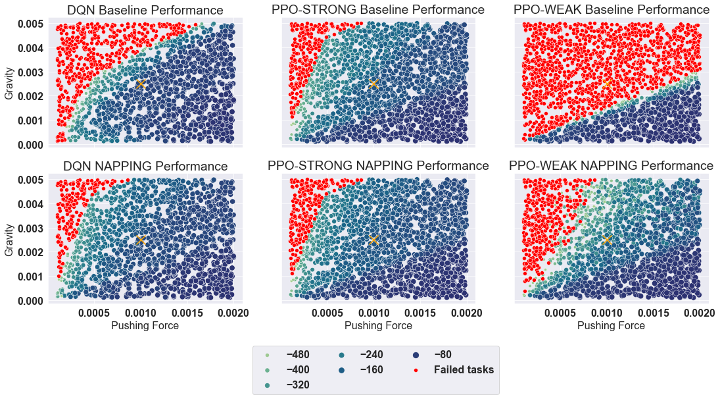}
\caption{Overall Cumulative Rewards on MountainCar by Novelties}
\label{fig:mountaincar_bymodel}
\end{figure}

Figure \ref{fig:mountaincar_bymodel} compares performance under different gravity and force values for all the -NAPPING agents. Each dot represents an evaluated trial with novel environment parameter values, with the $x$ axis showing the value of force and $y$ showing gravity. The color of the dots represents the average accumulative reward achieved in the environment. The larger the reward, the deeper the color and the larger the size of the dot. The first row/column shows the performance of the baseline model, while the second shows the performance of the same DRL agent but with \textit{NAPPING}. The red dots represent the environment where the agent fails, i.e., having -500 rewards. In general, the dots are darker in the bottom left direction and are lighter and eventually turn to red in the top left direction. The novel task becomes easier to solve with a larger pushing force value and a smaller gravity level. While the difficulty of the environment increases quickly in the top left direction, resulting in possibly unsolvable environments around the top left corner. The orange cross is the default parameter value of the domain. From the figure, we can see that, with \textit{NAPPING}, the proportion of failed tasks reduced significantly for both the PPO-Weak and DQN agents, witnessing 59.4\% and 71.45\% reduction in the number of failed tasks, respectively. Moreover, one can see that the performances of PPO-Strong-NAPPING and DQN-NAPPING are very similar. It indicates that with NAPPING, the DQN agent is able to reach the empirical maximum performance level. 

\subsection{CrossRoad}
CrossRoad is a toy domain inspired by OpenAI Freeway. Compared to OpenAI Freeway, CrossRoad allows the user to fully control the speeds and locations of moving objects and it allows the agent to move not just in upward and downward directions, but also in left and right directions. It has 8 "cars" (red boxes) moving horizontally at different speeds and directions. The goal is to cross the road while navigating the "player" (green box) and avoiding colliding with the cars. The agent can either move left, right, up, down, or stay in a place. The episode starts with the player being placed on the bottom line and ends if the player collides with a car, crosses the road, or reaches the episode length of 100 steps. A +1 reward is given if the agent crosses the road, -1, if the agent collides with a car or 100 steps are passed. This work assumes that the agent only has partial observable state space; the agent sees only its location and the adjacent eight cells, each with the $x$ and $y$ coordinates. Therefore, the state is an $18$ dimensional vector. The reward agent can get in each episode is therefore either -1 or 1.

\subsubsection{Baseline Agents}
As we do not have pre-trained agents for CrossRoad, we train a DQN and a PPO agent using the default setting on \cite{stable-baselines3}. The trained agents perform perfectly in the pre-novelty environment. Like previous domains, we run the offline (-Baseline) version, online learning (-Online), and fine-tuning (-FineTune) version and compare the performance with \textit{NAPPING} (-NAPPING) for both the DQN and PPO agent.

\subsubsection{Novelty}
We create eight novelty bases in CrossRoad, representing eight different novel layouts of the game environment. These bases are:

\begin{itemize}
\item Super Slow Speeds: the cars are still moving in the same direction, but the speed is much slower.
\item Super Fast Speeds: the cars are still moving in the same direction, but the speed is much faster.
\item New Speeds: the cars are moving in the same direction but at different speeds.
\item Opposite Direction: the cars are moving at the same speeds but in opposite directions.
\item All to the Left: All cars are moving to the left.
\item All to the Right: All cars are moving to the right.
\item Shift Speeds: speeds of the cars are shifted.
\item Reverse Cars: the order of the cars is reversed. i.e. the first car from the top now become the last car at the bottom.
\end{itemize}

We then further vary the initial position and the speed of cars for each novelty base by adding noise on both the initial position and the speed of cars. For each car's initial position, we add a uniformly sampled noise from $[-1, 1]$, and for speed, the noise is sampled from $[-10, 10]$. We also add this noise to the pre-novelty environment. In total, thus, we have nine different novelties. A novelty is detected if the five-rolling average episodic reward is less than 1.
We run 100 trials per novelty per DRL agent, totaling 1800 trials.

\subsubsection{CrossRoad Results and Discussion}

\begin{figure}[h]
\centering
\includegraphics[width=7cm, height=7cm]{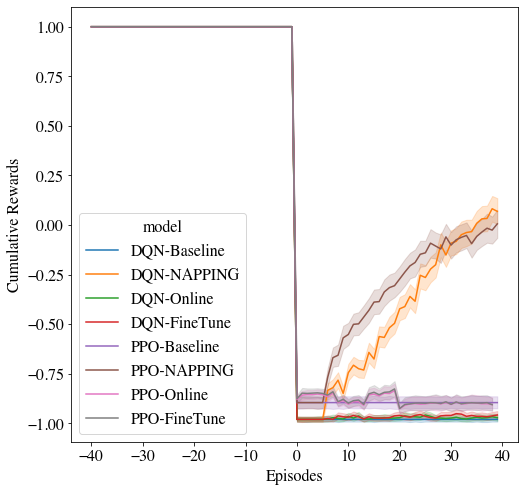}
\caption{Overall Cumulative Rewards on CrossRoad}
\label{fig:crossroad_overall}
\end{figure}

Figure \ref{fig:crossroad_overall} shows the overall performance of the evaluated agents. Similar to the results in previous domains, only agents with NAPPING showed adaptation behavior, with the DQN-NAPPING and PPO-NAPPING reaching around 0 cumulative rewards, indicating a 50\% of solving rate of novel tasks. In comparison, all other agents fail to learn any helpful policy to react to the novelties. When comparing PPO-NAPPING and DQN-NAPPING, we can see that although PPO-NAPPING has a slightly more robust baseline model and adapts faster than DQN-NAPPING, the asymptotic performance of PPO-NAPPING is slightly lower than that of DQN-NAPPING.

\begin{figure}[h]
\centering
\includegraphics[width=10cm, height=8cm]{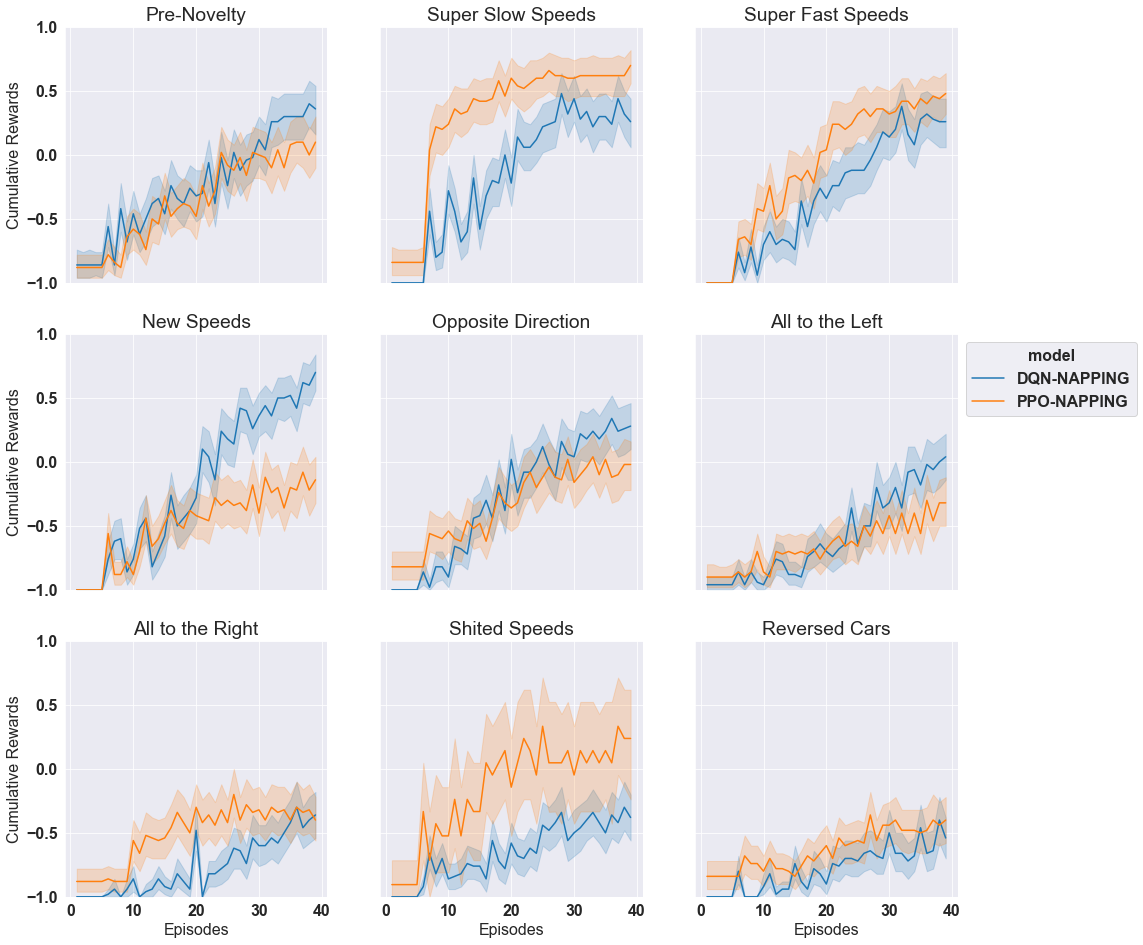}
\caption{Overall Cumulative Rewards on CrossRoad by Novelty}
\label{fig:crossroad_prenovelty}
\end{figure}

Figure \ref{fig:crossroad_prenovelty} displays the performance of the DRL agents with \textit{NAPPING}. Interestingly, different DRL agents with different embedded spaces show different adaptation performances. For example, DQN-NAPPING responds to the New Speed novelty much faster than PPO-NAPPING and has a much higher asymptotic performance. While PPO-NAPPING performs slightly better in All to the Right and Super Fast Speeds novelties and much better in Shifted Speeds and Super Slow Speeds novelties.

\subsection{Angry Birds}

Angry Birds, a simple and intuitive game with realistic physical simulation, has been one of the most popular testing domains for physical reasoning among the AI community \cite{renz2019ai, Xue2021}. The goal of Angry Birds is to destroy all pigs at a game level by shooting birds from a slingshot. Pigs are normally protected by physical structures made of blocks with varied sizes, shapes, and materials. Some birds have special powers that can be activated after being released from the slingshot. The only actions available to the players are to select a bird trajectory by pulling the bird back in the slingshot to the release coordinates (x, y) and then tapping the screen at time t after release to activate the special power. 

The AIBIRDS competition \cite{AIBirds} has been organized annually since 2012, mostly collocated with the International Joint Conference on Artificial Intelligence. It encourages the AI community to develop agents that can deal with continuous action space and environments in that the agent does not have complete knowledge about the physical parameters of objects. From 2021, the AIBIRDS Novelty Track has been introduced based on \cite{Xue2022} with a focus on promoting the development of agents that can handle unseen novelties in the physical world using physical reasoning skills. In this work, we evaluate our method using the novelties used in the most recent AIBIRDS Novelty Track competition. 

In the AIBIRDS Novelty Track competition, an agent can request two types of state representations: the screenshot and the ground truth. A screenshot state representation is a 480 x 640 colored image, and the ground truth representation is in JSON format containing all foreground objects in a screenshot. Each object in the ground truth representation is represented as a polygon of its vertices (provided in order) and its respective color map containing a list of 8-bit quantized colors that appear in the game object with their respective percentages. An agent can request screenshots or ground truth representations of the game level at any time while playing.

Since \textit{NAPPING} currently works only with agents with discrete action space, we discretize the action space of Angry Birds to three discrete actions: objects to shoot, high trajectory or low trajectory, and target offsets. More specifically, using the trajectory planner provided in the AIBIRDS Competition Novelty Track, the agent first decides which object to shoot at. Next, depending on whether the bird should reach the object from the top left side or just the left side, the agent needs to decide whether the high or low trajectory should be used. To increase the freedom of the agent's strategy and compensate for the noise in the trajectory planner, we also allow the agent to choose an offset for the target object from 7 different locations starting from no changes to 15/25/35 pixels vertical variations. We further simplify the action space by assuming the agent uses only the full strength of the slingshot, i.e. the bird is pulled to the furthermost point from the slingshot before being released. When \textit{NAPPING} sampling the adaptation action in Angry Birds, it always starts with the (object, trajectory, offset) triplet purposed by the baseline agent and searches for a triplet that with a maximum $Eval$ value by firstly varying objects, then trajectory and finally offset. 
\subsubsection{Baseline Agents}

For the Angry Birds domain, same as the DRL agents used in \cite{Xue2021}, we train a DQN \cite{wang2016dueling, van2016deep} agent and the DQN-Rel agent that contains a relational module \cite{zambaldi2018relational} to evaluate the performance of NAPPING. Both agents are trained on a set of pre-novelty tasks provided by the competition organizer following the same setup used in \cite{Xue2021}. The pre-novelty task set contains 2450 game levels that are generated from seven task templates. Both agents trained on in total of $20,000$ episodes with over $95\%$ of training pass rates. We compare our approach with a heuristic adaptation agent \textit{Naive Adaptation}, and two current state-of-the-art open world learning agents \textit{HYDRA} \cite{sternmodel,klenk2020model} and \textit{OpenMIND} \cite{musliner2021openmind}. \textit{OpenMIND} was the official champion of the 2022 AIBIRDS Competition Novelty Track. The \textit{Naive Adaptation} is built on top of the \textit{Naive} Agent, which shoots only at the pigs. \textit{Naive Adaptation} uses the strategy of the \textit{Naive} Agent in the pre-novelty game levels. After receiving the novelty signal, it searches for a combination of (objects, trajectories, and delays) that solve a game level and keeps a record of each triplet tried. Once a solution triplet (e.g., a solution triplet can be (pig\_1, high trajectory, delay 5 seconds)) is found for a trial, the \textit{Naive Adaptation} will keep using the triplet until it doesn't work anymore, where the agent starts to search for another triplet. We also include another Anonymous open-world learning agent and a non-adaptation agent \textit{DataLab} (the 2014 and 2015 AIBIRDS competition winner \cite{renz2016angry}) to examine the effect of the novelties on standard agents for comparison.

\subsubsection{Novelty}
There are sever novelties available in the 2022 AIBIRDS competition Novelty Track.

\begin{itemize}
\item Green Egg: A green rectangular object that has the same color as pigs but a different shape from the pigs
\item Bone: An object that needs to be destroyed to pass the game level. 
\item Taller Slingshot: Slingshot is now  twice high as normal - agent needs to adjust the release point accordingly
\item Pig Color: All the objects' colors changed to pig's color
\item Cicumcircle: circumcircle of objects instead of shapes in the ground truth
\item Floating Wood: Wood objects can float on the ice objects
\item Bird Likes Pig: Red Bird cannot cause damage to pigs
\end{itemize}

We evaluate all the agents following the trial setting, but unlike in previous domains, the number of pre-novelty episodes is a random number between each trial as that was the setting used in the competition. There are 40 novel episodes, and we run 40 trials for each novelty. 

\subsubsection{Angry Birds Results and Discussion}

\begin{figure}[h]
\centering
\includegraphics[width=13cm, height=8cm]{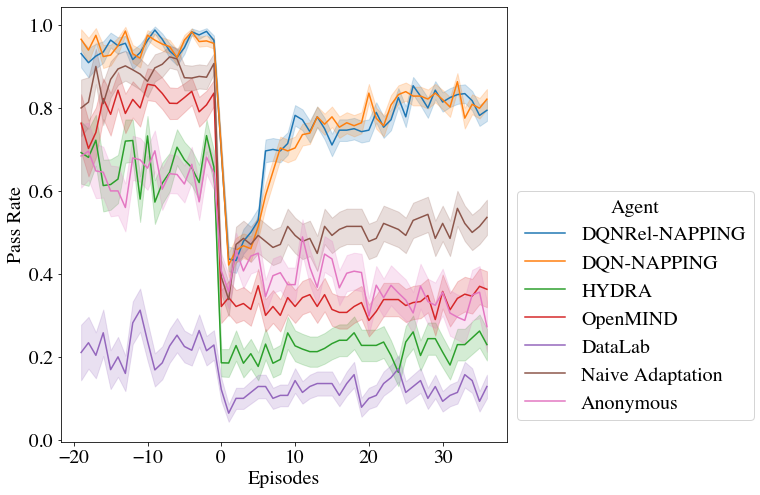}
\caption{Average pass rate of Agents on AIBIRDS Competition Novelty Track by Agent}
\label{fig:aibirds_overall}
\end{figure}

Figure \ref{fig:aibirds_overall} demonstrates the overall pass rate of the seven agents tested in the Angry Birds domain. Unlike in the previous domains, in Angry Birds the number of pre-novelty episodes is random; therefore, we set the episode where novelty is introduced to be $0$, and pre-novelty episodes are, therefore, the negative ones. \textit{DQN-NAPPING} and \textit{DQNRel-NAPPING} have the strongest novelty adaptation performance among all the agents evaluated. Averaging just above 0.9 in the pre-novelty instances, the pass rates of the -NAPPING agents drop to below 0.4 right after a novelty is presented and then quickly recover to around 0.8 before the end of the trial. The \textit{Naive Adaptation} agent, although it responds quickly to some of the novel tasks in just 1 episode, has a lower asymptotic performance compared to that of \textit{DQN-NAPPING} and \textit{DQNRel-NAPPING}.  \textit{OpenMIND}, as one of the state-of-the-art open-world learning agents, achieved above 0.8 pass rate in the pre-novelty and above 0.4 in the post-novelty environment. It is interesting to notice that both \textit{HYDRA} and the \textit{Anonymous} agents have similar pre-novelty performance, \textit{Anonymous} is more novelty resilient than \textit{HYDRA} with a less significant reduction in the performance after a novelty is introduced. However, \textit{Anonymous} seemed to struggle to find the strategy to adapt to novelties, as the average pass rate gradually drops after novelty is introduced. Despite being the previous champion in the standard AIBIRDS Competition track, the non-adapting \textit{DataLab} has the lowest overall performance among all the agents, with a pre-novelty pass rate of around 0.3 to below 0.2 in the post-novelty episodes. 

\begin{figure}[h]
\centering
\includegraphics[width=12cm, height=10cm]{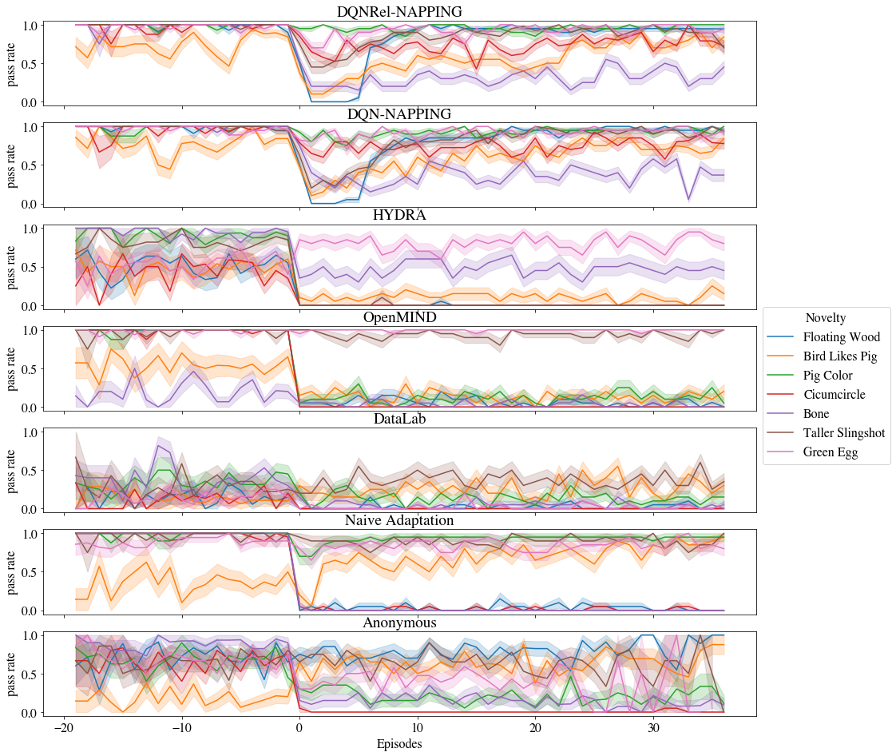}
\caption{Average pass rate of Agents on AIBIRDS Competition Novelty Track by Novelty}
\label{fig:airbirds_byagent}
\end{figure}

Figure \ref{fig:airbirds_byagent} shows the pass rate of the agents by different novelties. We can see that the -NAPPING agents respond to all novelties, returning to the pre-novelty performance except for the Bone novelty, where the agents need to understand the new object Bone is another type of pig and needs to be destroyed to pass the game level. The -NAPPING agents are also robust to the Pig Color novelty, as the agents (also the \textit{Naive Adaptation} agent) use the object's label directly instead of the color maps. Nevertheless, the \textit{Naive Adaptation} agent fails to adapt to the Cicumcircle and the Floating Wood novelty. For the Cicumcircle novelty, it is because the novelty breaks the \textit{Naive Adaptation}'s input state assumption that objects are represented by vertices and hence causes a crash. The reason for \textit{Naive Adaptation} not reacting to the Floating Wood novelty is that an agent needs to use different trajectories accordingly in different game levels, whereas \textit{Naive Adaptation} can remember only one solution triplets. For example, a high trajectory must be used when objects are closer to the slingshot, and a lower trajectory is required if the high trajectory path is blocked and when the game objects are far away from the slingshot. On the other hand, the -NAPPING agents do not have particular assumptions on the input states as \textit{NAPPING} works directly over the embedding space of DRL models. Also, \textit{NAPPING} agents can identify the different solution actions for tasks with different spatial arrangements and timing requirements, as these differences are assumed to have different representations in the embedded space of the DRL agent. We include the agents' pass rate on each novelty in \ref{sec:Performance:AIBIRDSnovelty}.

\begin{figure}[h]
\centering
\includegraphics[width=12cm, height=2.5cm]{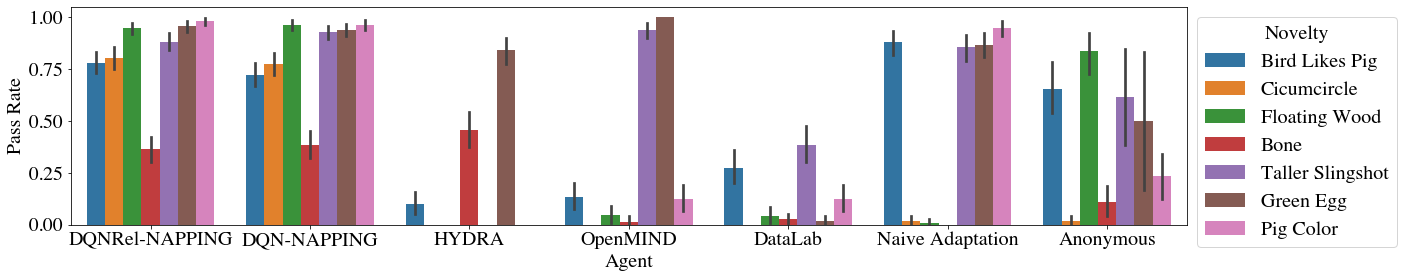}
\caption{Average asymptotic (last 10 episodes) pass rate on AIBIRDS Competition Novelty Track}
\label{fig:aibirds_asymptoticl}
\end{figure}

Figure \ref{fig:aibirds_asymptoticl} shows the asymptotic (last ten episodes) pass rate of different agents and for different novelties. The -NAPPING agents have similar asymptotic performance levels, with DQNRel being slightly better in the Bird Likes Pig and Cicumcicle novelty and slightly worse in other novelties. The \textit{Naive Adaptation} has a relatively high pass rate if the heuristic it has can solve the novelties, and it fails if the heuristic does not work. \textit{OpenMIND} is particularly good in adapting the Green Egg and Taller Slingshot novelty, achieving the highest asymptotic pass rate among all agents but did not perform well in other novelties. The \textit{Anonymous} agent seems to be very robust, with all pass rates higher than the non-adapting agent \textit{DataLab}.

\section{Conclusion and Future Work}
\label{sec:conclusion}

In this work, we propose \textit{NAPPING}, a learning method that works with DRL agents to address the open-world learning challenge. We evaluate our method with different DRL learning agents with various performance levels in the pre-novelty environment in four domains with different tasks and rewards. Our results show that \textit{NAPPING} can quickly adapt to a novel situation. Our method outperforms the standard machine learning approaches (Online learning and fine-tuning) and the current state-of-the-art open-world learning agents by a large margin.

\paragraph{Limitations and Future Work} many potential improvements can make \textit{NAPPING} more general. One of the future works may focus on alleviating the need for the pre-define $Eval$ function and the threshold $thre$. Our current implementation of $Eval$ and $thre$ seeks a "greedy" adaptation strategy as \textit{NAPPING} always maximize the $Eval$ value for the current action. But $Eval$ evaluates only the $s_t$ and $s_{t+1}$, and it does not reflect the value of action in the long run. One way to address this issue is to adequately update the $Eval$ function once the agent enters the novel world by using the model states and the episodic rewards. Also, extending \textit{NAPPING} to work for agents with continuous action space would be helpful for \textit{NAPPING} to work on more environments. Last but not least, instead of sampling actions from the action space, learning what action should be selected more likely to achieve the task would improve the efficiency of \textit{NAPPING}. 

\section*{Acknowledgments}

This research was sponsored by the Defense Advanced Research Projects Agency (DARPA) and the Army Research Office (ARO) and was accomplished under Cooperative Agreement Number W911NF-20-2-0002. The views and conclusions contained in this document are those of the authors and should not be interpreted as representing the official policies, either expressed or implied, of the DARPA or ARO, or the U.S. Government. The U.S. Government is authorized to reproduce and distribute reprints for Government purposes notwithstanding any copyright notation herein.

\bibliographystyle{elsarticle-num} 
\bibliography{cas-refs}

\begin{thebibliography}{10}
\expandafter\ifx\csname url\endcsname\relax
  \def\url#1{\texttt{#1}}\fi
\expandafter\ifx\csname urlprefix\endcsname\relax\def\urlprefix{URL }\fi
\expandafter\ifx\csname href\endcsname\relax
  \def\href#1#2{#2} \def\path#1{#1}\fi

\bibitem{arulkumaran2017brief}
K.~Arulkumaran, M.~P. Deisenroth, M.~Brundage, A.~A. Bharath, A brief survey of
  deep reinforcement learning, arXiv preprint arXiv:1708.05866 (2017).

\bibitem{Vinyals2019}
O.~Vinyals, I.~Babuschkin, W.~M. Czarnecki, M.~Mathieu, A.~Dudzik, J.~Chung,
  D.~H. Choi, R.~Powell, T.~Ewalds, P.~Georgiev, et~al., Grandmaster level in
  starcraft ii using multi-agent reinforcement learning, Nature 575~(7782)
  (2019) 350--354.

\bibitem{silver2016mastering}
D.~Silver, A.~Huang, C.~J. Maddison, A.~Guez, L.~Sifre, G.~Van Den~Driessche,
  J.~Schrittwieser, I.~Antonoglou, V.~Panneershelvam, M.~Lanctot, et~al.,
  Mastering the game of go with deep neural networks and tree search, nature
  529~(7587) (2016) 484--489.

\bibitem{lazaridis2020deep}
A.~Lazaridis, A.~Fachantidis, I.~Vlahavas, Deep reinforcement learning: A
  state-of-the-art walkthrough, Journal of Artificial Intelligence Research 69
  (2020) 1421--1471.

\bibitem{witty2021measuring}
S.~Witty, J.~K. Lee, E.~Tosch, A.~Atrey, K.~Clary, M.~L. Littman, D.~Jensen,
  Measuring and characterizing generalization in deep reinforcement learning,
  Applied AI Letters 2~(4) (2021) e45.

\bibitem{Xue2022}
C.~Xue, V.~Pinto, P.~Zhang, C.~Gamage, E.~Nikonova, J.~Renz, Science birds
  novelty: An open-world learning test-bed for physics domains, Proceedings of
  the AAAI Conference on Artificial Intelligence, Designing Artificial
  Intelligence for Open Worlds (2022).

\bibitem{AIBirds}
AIBIRDS, \href{http://aibirds.org/}{Angry birds {AI} competition} [cited
  22.11.2022].
\newline\urlprefix\url{http://aibirds.org/}

\bibitem{Senator2019}
T.~Senator,
  \href{https://www.darpa.mil/program/science-of-artificial-intelligence-and-learning-for-open-world-novelty}{Science
  of artificial intelligence and learning for open-world novelty ({SAIL-ON})}
  (2019) [cited 10.11.2022].
\newline\urlprefix\url{https://www.darpa.mil/program/science-of-artificial-intelligence-and-learning-for-open-world-novelty}

\bibitem{Langley2020}
P.~Langley, Open-world learning for radically autonomous agents, in:
  Proceedings of the AAAI Conference on Artificial Intelligence, Vol.~34, 2020,
  pp. 13539--13543.

\bibitem{Langley2022}
P.~Langley, Constraints on theories of open-world learning, in: Proceedings of
  the AAAI Conference on Artificial Intelligence, Designing Artificial
  Intelligence for Open Worlds, 2022.

\bibitem{https://doi.org/10.48550/arxiv.2011.12906}
M.~Jafarzadeh, A.~R. Dhamija, S.~Cruz, C.~Li, T.~Ahmad, T.~E. Boult,
  \href{https://arxiv.org/abs/2011.12906}{A review of open-world learning and
  steps toward open-world learning without labels} (2020).
\newblock \href {https://doi.org/10.48550/ARXIV.2011.12906}
  {\path{doi:10.48550/ARXIV.2011.12906}}.
\newline\urlprefix\url{https://arxiv.org/abs/2011.12906}

\bibitem{Boult2021}
T.~Boult, P.~A. Grabowicz, D.~Prijatelj, R.~Stern, L.~Holder, J.~Alspector,
  M.~Jafarzadeh, T.~Ahmad, A.~R. Dhamija, Cli, S.~Cruz, A.~Shrivastava,
  C.~Vondrick, W.~Scheirer, Towards a unifying framework for formal theories of
  novelty, Proceedings of the AAAI Conference on Artificial Intelligence 35
  (2021) 15047--15052.

\bibitem{Molineaux2022}
M.~Molineaux, D.~Dannenhauer, An environment transformation-based framework for
  comparison of open-world learning agents, in: Proceedings of the AAAI
  Conference on Artificial Intelligence, Designing Artificial Intelligence for
  Open Worlds, 2022.

\bibitem{Liu2022AIAS}
B.~Liu, S.~Mazumder, E.~Robertson, S.~Grigsby, Ai autonomy: Self-initiation,
  adaptation and continual learning, ArXiv abs/2203.08994 (2022).

\bibitem{muhammad2021novelty}
F.~Muhammad, V.~Sarathy, G.~Tatiya, S.~Goel, S.~Gyawali, M.~Guaman, J.~Sinapov,
  M.~Scheutz, A novelty-centric agent architecture for changing worlds, in:
  Proceedings of the 20th International Conference on Autonomous Agents and
  MultiAgent Systems, 2021, pp. 925--933.

\bibitem{burachasmetacognitive}
G.~T. Burachas, S.~Grigsby, W.~Ferguson, J.~Krichmar, R.~Rao, Metacognitive
  mechanisms for novelty processing: Lessons for ai.

\bibitem{goel2022rapid}
S.~Goel, Y.~Shukla, V.~Sarathy, M.~Scheutz, J.~Sinapov, Rapid-learn: A
  framework for learning to recover for handling novelties in open-world
  environments, arXiv preprint arXiv:2206.12493 (2022).

\bibitem{NovGrid}
J.~Balloch, Z.~Lin, M.~Hussain, A.~Srinivas, R.~Wright, X.~Peng, J.~Kim,
  M.~Riedl, Novgrid: A flexible grid world for evaluating agent response to
  novelty, arXiv preprint arXiv:2203.12117 (2022).

\bibitem{khetarpal2020towards}
K.~Khetarpal, M.~Riemer, I.~Rish, D.~Precup, Towards continual reinforcement
  learning: A review and perspectives, arXiv preprint arXiv:2012.13490 (2020).

\bibitem{padakandla2020reinforcement}
S.~Padakandla, P.~KJ, S.~Bhatnagar, Reinforcement learning algorithm for
  non-stationary environments, Applied Intelligence 50~(11) (2020) 3590--3606.

\bibitem{cheung2020reinforcement}
W.~C. Cheung, D.~Simchi-Levi, R.~Zhu, Reinforcement learning for non-stationary
  markov decision processes: The blessing of (more) optimism, in: International
  Conference on Machine Learning, PMLR, 2020, pp. 1843--1854.

\bibitem{yu2020meta}
T.~Yu, D.~Quillen, Z.~He, R.~Julian, K.~Hausman, C.~Finn, S.~Levine,
  Meta-world: A benchmark and evaluation for multi-task and meta reinforcement
  learning, in: Conference on robot learning, PMLR, 2020, pp. 1094--1100.

\bibitem{taylor2009transfer}
M.~E. Taylor, P.~Stone, Transfer learning for reinforcement learning domains: A
  survey., Journal of Machine Learning Research 10~(7) (2009).

\bibitem{zhu2020transfer}
Z.~Zhu, K.~Lin, J.~Zhou, Transfer learning in deep reinforcement learning: A
  survey, arXiv preprint arXiv:2009.07888 (2020).

\bibitem{fernandez2006probabilistic}
F.~Fern{\'a}ndez, M.~Veloso, Probabilistic policy reuse in a reinforcement
  learning agent, in: Proceedings of the fifth international joint conference
  on Autonomous agents and multiagent systems, 2006, pp. 720--727.

\bibitem{barreto2017successor}
A.~Barreto, W.~Dabney, R.~Munos, J.~J. Hunt, T.~Schaul, H.~P. van Hasselt,
  D.~Silver, Successor features for transfer in reinforcement learning,
  Advances in neural information processing systems 30 (2017).

\bibitem{mclure2022changepoint}
M.~D. McLure, D.~J. Musliner, A changepoint method for open-world novelty
  detection, in: IGARSS 2022-2022 IEEE International Geoscience and Remote
  Sensing Symposium, IEEE, 2022, pp. 5329--5332.

\bibitem{boult2022weibull}
T.~E. Boult, N.~M. Windesheim, S.~Zhou, C.~Pereyda, L.~B. Holder,
  Weibull-open-world (wow) multi-type novelty detection in cartpole3d,
  Algorithms 15~(10) (2022) 381.

\bibitem{li2021unsupervised}
R.~Li, H.~Hua, P.~Haslum, J.~Renz, Unsupervised novelty characterization in
  physical environments using qualitative spatial relations, in: Proceedings of
  the International Conference on Principles of Knowledge Representation and
  Reasoning, Vol.~18, 2021, pp. 454--464.

\bibitem{kumar2021rma}
A.~Kumar, Z.~Fu, D.~Pathak, J.~Malik, Rma: Rapid motor adaptation for legged
  robots, arXiv preprint arXiv:2107.04034 (2021).

\bibitem{sternmodel}
R.~Stern, W.~Piotrowski, M.~Klenk, J.~de~Kleer, A.~Perez, J.~Le, S.~Mohan,
  Model-based adaptation to novelty in open-world ai, Proceedings of the 32nd
  International Conference on Automated Planning and Scheduling, Bridging the
  Gap Between AI Planning and Reinforcement Learning (2022).

\bibitem{klenk2020model}
M.~Klenk, W.~Piotrowski, R.~Stern, S.~Mohan, J.~de~Kleer, Model-based novelty
  adaptation for open-world ai, in: International Workshop on Principles of
  Diagnosis (DX), 2020.

\bibitem{musliner2021openmind}
D.~J. Musliner, M.~J. Pelican, M.~McLure, S.~Johnston, R.~G. Freedman,
  C.~Knutson, Openmind: Planning and adapting in domains with novelty,
  Proceedings of the Ninth Annual Conference on Advances in Cognitive Systems
  (2021).

\bibitem{aurenhammer1991voronoi}
F.~Aurenhammer, Voronoi diagrams—a survey of a fundamental geometric data
  structure, ACM Computing Surveys (CSUR) 23~(3) (1991) 345--405.

\bibitem{Barto1983NeuronlikeAE}
A.~G. Barto, R.~S. Sutton, C.~W. Anderson, Neuronlike adaptive elements that
  can solve difficult learning control problems, IEEE Transactions on Systems,
  Man, and Cybernetics SMC-13 (1983) 834--846.

\bibitem{Brockman2016OpenAIG}
G.~Brockman, V.~Cheung, L.~Pettersson, J.~Schneider, J.~Schulman, J.~Tang,
  W.~Zaremba, Openai gym, ArXiv abs/1606.01540 (2016).

\bibitem{Moore90efficientmemory-based}
A.~W. Moore, Efficient memory-based learning for robot control, Tech. rep.,
  University of Cambridge (1990).

\bibitem{CrossroadDomain}
nikonkate, Novelty domains, \url{https://github.com/nikonkate/novelty-domains}
  (2022).

\bibitem{Pinto2022}
V.~Pinto, J.~Renz, C.~Xue, P.~Zhang, K.~Doctor, D.~W. Aha, Measuring the
  performance of open-world {AI} systems, Proceedings of the AAAI Conference on
  Artificial Intelligence, Designing Artificial Intelligence for Open Worlds
  (2022).

\bibitem{stable-baselines3}
A.~Raffin, A.~Hill, A.~Gleave, A.~Kanervisto, M.~Ernestus, N.~Dormann,
  \href{http://jmlr.org/papers/v22/20-1364.html}{Stable-baselines3: Reliable
  reinforcement learning implementations}, Journal of Machine Learning Research
  22~(268) (2021) 1--8.
\newline\urlprefix\url{http://jmlr.org/papers/v22/20-1364.html}

\bibitem{renz2019ai}
J.~Renz, X.~Ge, M.~Stephenson, P.~Zhang, Ai meets angry birds, Nature Machine
  Intelligence 1~(7) (2019) 328--328.

\bibitem{Xue2021}
C.~Xue*, V.~Pinto*, C.~Gamage*, E.~Nikonova, P.~Zhang, J.~Renz, Phy-q as a
  measure for physical reasoning intelligence., Nature Machine Intelligence 5
  (2023) 83--93, *equal contribution.

\bibitem{wang2016dueling}
Z.~Wang, T.~Schaul, M.~Hessel, H.~Hasselt, M.~Lanctot, N.~Freitas, Dueling
  network architectures for deep reinforcement learning, in: International
  conference on machine learning, PMLR, 2016, pp. 1995--2003.

\bibitem{van2016deep}
H.~Van~Hasselt, A.~Guez, D.~Silver, Deep reinforcement learning with double
  q-learning, in: Proceedings of the AAAI conference on artificial
  intelligence, Vol.~30, 2016.

\bibitem{zambaldi2018relational}
V.~Zambaldi, D.~Raposo, A.~Santoro, V.~Bapst, Y.~Li, I.~Babuschkin, K.~Tuyls,
  D.~Reichert, T.~Lillicrap, E.~Lockhart, et~al., Relational deep reinforcement
  learning, arXiv preprint arXiv:1806.01830 (2018).

\bibitem{renz2016angry}
J.~Renz, X.~Ge, R.~Verma, P.~Zhang, Angry birds as a challenge for artificial
  intelligence, in: Proceedings of the AAAI Conference on Artificial
  Intelligence, Vol.~30, 2016.

\end{thebibliography}
\newpage

\appendix

\section{Performance of all agents per novelty in the AIBIRDS competition novelty track}
\label{sec:Performance:AIBIRDSnovelty}

\begin{figure}[h]
\centering
\includegraphics[width=14cm, height=13.5cm]{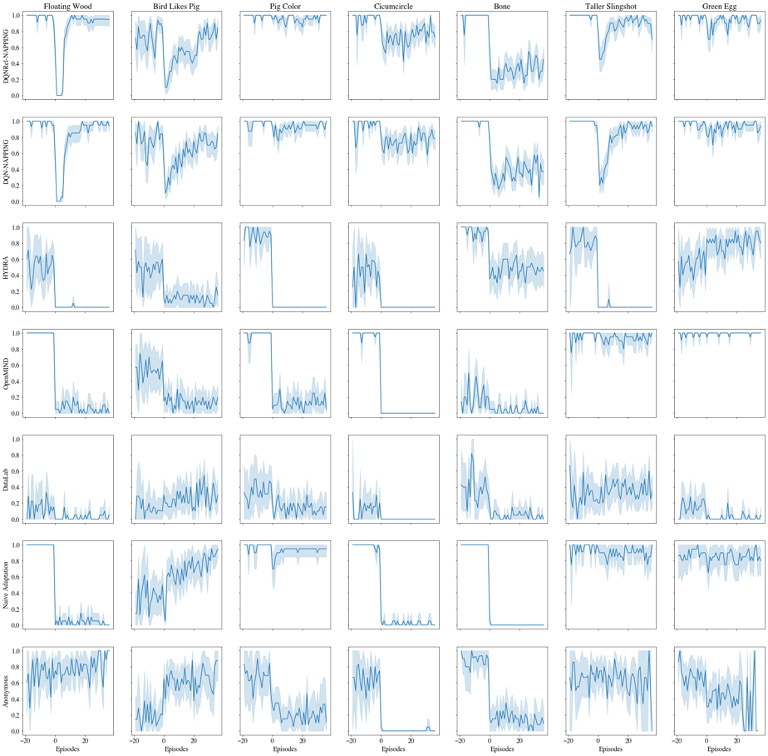}
\caption{Performance of all agents per novelty in the AIBIRDS competition novelty track}
\label{fig:aibirds_pernovelty}
\end{figure}

\newpage

\section{Algorithm that selects an action from the Adaptation Principles}
\label{sec:algorithm:policy}
\begin{algorithm}[ht]
\scriptsize

\caption{$\pi_{AP}$: \textit{adaptation principle} Policy}\label{alg:piap}
\begin{algorithmic}
\Require
\State $AdaptationPrinciples$
\State $VoronoiSet$

\Inputs{
\State $ms_{t}$ : model state of the agent at time $t$
\State $a^{agent}_{t}$: action of the agent at time $t$
}

\If {$size(AdaptationPrinciples) = 0$}
    \State $a^{ap}_{t} = a^{agent}_{t}$
\Else
    \State $ms_{region} = $ the closet point in $VoronoiSet$ to $ms_t$
    \If{$(ms_{region},a^{agent}_{t})$ in $AdaptationPrinciples$}
        \State $PrincipleActions = AdaptationPrinciples[(ms_{region},a^{agent}_{t})]$
        \If{$size(PrincipleActions) > 1$} \Comment{\textit{adaptation principle} is learning}
            \State $a^{ap}_{t} =$ Choose an action from $PrincipleActions$ randomly

        \Else \Comment{\textit{adaptation principle} is learned}
            \State $a^{ap}_{t} = PrincipleActions$
    \EndIf
    \Else
        \State $a^{ap}_{t} = a^{agent}_{t}$
    \EndIf
\EndIf

\State Return $a^{ap}_{t}$

\end{algorithmic}
\end{algorithm}

\section{Algorithm that updates the Adaptation Principles}
\label{sec:algorithm:update}
\begin{algorithm}[ht]
\tiny
\caption{$\phi_{update}$: updating the \textit{adaptation principle} policy}\label{alg:update}
\begin{algorithmic}

\Require
\State $AdaptationPrinciples$
\State $VoronoiSet$
\State $BestScore$
\State $Eval: S \times A \times S \rightarrow \mathbb{R}$: function that evaluates the score of an action
\State $thre$: a threshold that decides if a new principle needs to be created
\State $A$: set of available actions in the environment

\Inputs{
\State $ms_{t}$ : model state of the agent at time $t$
\State $a^{agent}_{t}$: action of the agent at time $t$
\State $a^{ap}_{t}$: action of $\pi_{AP}$ at time $t$
\State $s_{t}$ : environment state at time step $t$
\State $s_{t+1}$ : environment state at time step $t+1$
}
\\

\State $actionScore = Eval(s_{t}, a_t, s_{t+1})$
\If {$size(AdaptationPrinciples) = 0$}
    \If {$actionScore >= thre $}
    \State $AdaptationPrinciples[(ms_t, a^{agent}_{t})] = \{a^{agent}_{t}\}$
    \Else
    \State $AdaptationPrinciples[(ms_t, a^{agent}_{t})] = \{a \in A\} - \{a^{agent}_{t}\}$
    \EndIf
    \State $BestScore[(ms_{t},a^{agent}_{t})] = actionScore$
    \State add $ms_t$ to $VoronoiSet$
\Else
    \State $ms_{region} = $ the closet point in $VoronoiSet$ to $ms_t$
    \If {$(ms_{region}, a^{agent}_{t})$ in $AdaptationPrinciples$}
        \State $PrincipleActions = AdaptationPrinciples[(ms_t, a^{agent}_{t})]$
        \State $BestRegionScore = BestScore[(ms_{region},a^{agent}_{t})]$

            \State \textbf{Case} {$actionScore == sup(Eval(\cdot)):$} 
            \If { $size(PrincipleActions) > 1$}
                \State Remove all other actions in $PrincipleActions$
                \State $BestScore[(ms_{region},a^{agent}_{t})] = actionScore$
            \EndIf
            
            \State \textbf{Case} {$actionScore > BestRegionScore$:} 
            \If { $size(PrincipleActions) > 1$}
                \State Remove the other tested actions in $PrincipleActions$
                \State $BestScore[(ms_{region},a^{agent}_{t})] = actionScore$
            \EndIf
    
            \State \textbf{Case} {$actionScore == BestRegionScore$ and $actionScore \geq thre$:}
            \If { $size(PrincipleAction) > 1$}
                \State Remove the other tested actions in $PrincipleActions$
                \State $BestScore[(ms_{region},a^{agent}_{t})] = actionScore$
            \EndIf
            
            \State \textbf{Case} {$actionScore == BestRegionScore$ and $actionScore < thre$:}
            \If { $size(PrincipleAction) > 1$} 
                \State Remove the action $a^{ap}_{t}$ from $PrincipleActions$
            \EndIf     
        
            \State \textbf{Case} {$actionScore < BestRegionScore$ and $actionScore < thre$:}
            \If { $size(PrincipleAction) > 1$} 
                \State Remove the action $a^{ap}_{t}$ from $PrincipleActions$
            \Else
                \State $AdaptationPrinciples[(ms_t, a^{agent}_{t})] = \{a \in \mathbb{A}\} - \{a^{agent}_{t}\}$
                \State $BestScore[(ms_{t},a^{agent}_{t})] = actionScore$
                \State add $ms_t$ to $VoronoiSet$
            \EndIf     
    \Else 
        \If{$actionScore \geq thre$} 
            $AdaptationPrinciples[(ms_t, a^{agent}_{t})] = \{a^{agent}_{t}\}$
        \Else
        \State $AdaptationPrinciples[(ms_t, a^{agent}_{t})] = \{a \in \mathbb{A}\} - \{a^{agent}_{t}\}$
        \EndIf
        \State $BestScore[(ms_{t},a^{agent}_{t})] = actionScore$
        \State add $ms_t$ to $VoronoiSet$
    \EndIf
\EndIf

\end{algorithmic}
\end{algorithm}





% \end{thebibliography}
%\end{linenumbers}

\end{document}